%% file: acl_latex.tex
\pdfoutput=1

\documentclass[11pt]{article}

\usepackage[final]{acl}

\usepackage{times}
\usepackage{latexsym}

\usepackage[T1]{fontenc}

\usepackage[utf8]{inputenc}

\usepackage{microtype}

\usepackage{inconsolata}

\usepackage{graphicx}

\usepackage{soul}
\definecolor{lightblue}{rgb}{0.68, 0.85, 0.9}
\sethlcolor{lightblue}

\usepackage{amsmath,amssymb,amsthm,amsfonts}
\usepackage{dsfont}

\usepackage{geometry}
\usepackage{booktabs}
\usepackage{makecell}
\usepackage{multirow}
\usepackage{subcaption}

\usepackage[most]{tcolorbox}
\usepackage{fontawesome5}
\definecolor{yellow}{HTML}{F6BD60}
\usepackage{multicol}

\definecolor{rcolor}{RGB}{255,242,204}
\definecolor{ccolor}{RGB}{213,232,212}

\definecolor{hilight}{HTML}{000000}
\definecolor{hilight}{HTML}{FF0000}
\sethlcolor{hilight}

\usepackage{rotating}
\usepackage{tabularx}

\usepackage{algorithm}
\usepackage{algpseudocode}

\usepackage{colortbl}

\usepackage[most]{tcolorbox}

\usepackage{enumitem} %

\theoremstyle{remark}

\title{Two Heads Are Better Than One: Dual-Model Verbal Reflection at Inference-Time}

\author{Jiazheng Li$^1$\quad Yuxiang Zhou$^{1,6}$\quad Junru Lu$^4$\quad Gladys Tyen$^5$\thanks{Now at Google DeepMind.}\\
{\bf Lin Gui$^1$\quad Cesare Aloisi$^2$\quad Yulan He$^{1,3}$}\\ 
  $^1$King's College London\quad\quad$^2$AQA\quad\quad$^3$The Alan Turing Institute\\$^4$Tencent YouTu Lab\quad\quad $^5$University of Cambridge\\
  $^6$Queen Mary University of London\\
\texttt{caloisi@aqa.org.uk, junrulu@tencent.com, gladys.tyen@cl.cam.ac.uk},\\
\texttt{\{jiazheng.li, yuxiang.zhou, lin.gui, yulan.he\}@kcl.ac.uk}}

\begin{document}
\maketitle

\begin{abstract}
Although preference optimization methods have improved reasoning performance in Large Language Models (LLMs), they often lack transparency regarding why one reasoning outcome is preferred over another. This limitation is especially critical in Automated Student Answer Scoring (ASAS), where explainability is essential to justify assessment outcomes. Verbal reinforcement learning offers the potential to generate explicit reflection, but it tends to produce superficial critiques that can harm assessment performance. Existing LLMs also struggle to reliably detect subtle reasoning errors in ASAS tasks. Moreover, manually identifying intermediate reasoning errors is expensive and difficult to scale. To address these challenges, we introduce a \textbf{contrastive reflection synthesis pipeline} that generates precise verbal feedback by identifying discrepancies in structure reasoning graph paths. Leveraging these synthetic reflection data, we propose \texttt{DARS}, a Dual-model Reflective Scoring framework featuring a dedicated Critic model trained for effective reflection. \texttt{DARS} achieves strong performance and consistently outperforms existing ASAS baselines across all evaluation metrics. Extensive experiments further provide novel insights into the value of reflection data, framework design, and the scaling behavior of \texttt{DARS}.\footnote{We release the DARS code at \url{https://github.com/lijiazheng99/DARS}.}
\end{abstract}

\input{sections/1_introduction_update_GL}

\input{sections/3_method}
\input{sections/4_experiment}

\input{sections/2_related_work}

\section{Conclusion and Discussion}

We proposed a novel approach to enhance %
reasoning through a dual-model framework, and also introduced a contrastive reflection synthesis pipeline, which generates more targeted verbal reflections. %
Our framework, consisting of a dedicated Reasoner and Critic, enables effective reasoning refinement without relying on oracle labels. %
Moreover, our carefully designed training process equips both models with capabilities that extend beyond task-specific reasoning. The Reasoner not only solves problems but also learns to refine its reasoning based on feedback, while the Critic not only identifies errors but also learns when to stop, ensuring efficient reasoning improvement.

\section*{Limitations}
This study has several limitations. First, the training process requires substantial computational resources. While our framework minimizes the need for future retraining, the SFT training for both the Reasoner and Critic involves additional data points to enhance the model’s various capabilities, leading to higher training FLOPs than single Reasoner approaches. Second, the generalizability of our framework to tasks beyond ASAS remains unexplored. Although we conducted a comprehensive evaluation across six datasets, our focus was predominantly on the ASAS task. Future work should investigate the applicability of the proposed framework to a broader range of tasks. For instance, while math and code reasoning problems may not necessitate a binary structured thought-tree approach, they could benefit from pre-defined rules to verify the correctness of intermediate steps and then identify path discrepancies.
Finally, our prompt design was not exhaustively optimized. Future work could incorporate in-context learning~\cite{zhou-etal-2024-mystery} and chain-of-thought prompting~\cite{wei2022chain} to further improve performance.

\section*{Ethics Statement}
This study utilized both public and proprietary datasets of anonymized student responses, none of which contain sensitive or personally identifiable information. We thoroughly reviewed the LLMs' outputs and did not identify any instances of harmful content or exposure of personal information. Nevertheless, before deploying our framework in high-stakes examination settings, experts must carefully evaluate its assessment decisions and the underlying rationales to ensure reliability and fairness.

\section*{Acknowledgments}
This work was supported in part by the UK Engineering and Physical Sciences Research Council through a Turing AI Fellowship (grant no. EP/V020579/1, EP/V020579/2) and a Prosperity Partnership project with AQA (UKRI566). Jiazheng Li is funded by a PhD scholarship provided by AQA. We thank Hainiu Xu and Ruobing Wang for their advice on formatting for this paper.

\clearpage
\bibliography{custom}

\clearpage

\appendix
\setcounter{table}{0}
\renewcommand{\thetable}{A\arabic{table}}
\setcounter{figure}{0}
\renewcommand{\thefigure}{A\arabic{figure}}

\input{sections/5_experiment_setup}

\input{sections/7_further_exps}

\end{document}

%% file: sections/1_introduction_update_GL.tex
\section{Introduction}
Automated Student Answer Scoring (ASAS) is a crucial educational NLP task that aims to automate the intricate reasoning process performed by human graders. It offers the potential for faster and more consistent assessment at scale. To enhance transparency in automated decisions, recent studies have incorporated Large Language Models (LLMs) to generate free‑form rationales alongside scoring~\cite{li_emnlp2023, aaai_demo}. However, these generated rationales are often \emph{partially} correct, mixing valid logic with subtle yet impactful errors~\cite{li-etal-2025-drift}.

\begin{figure}[!t]
\centering
\includegraphics[width=\linewidth]{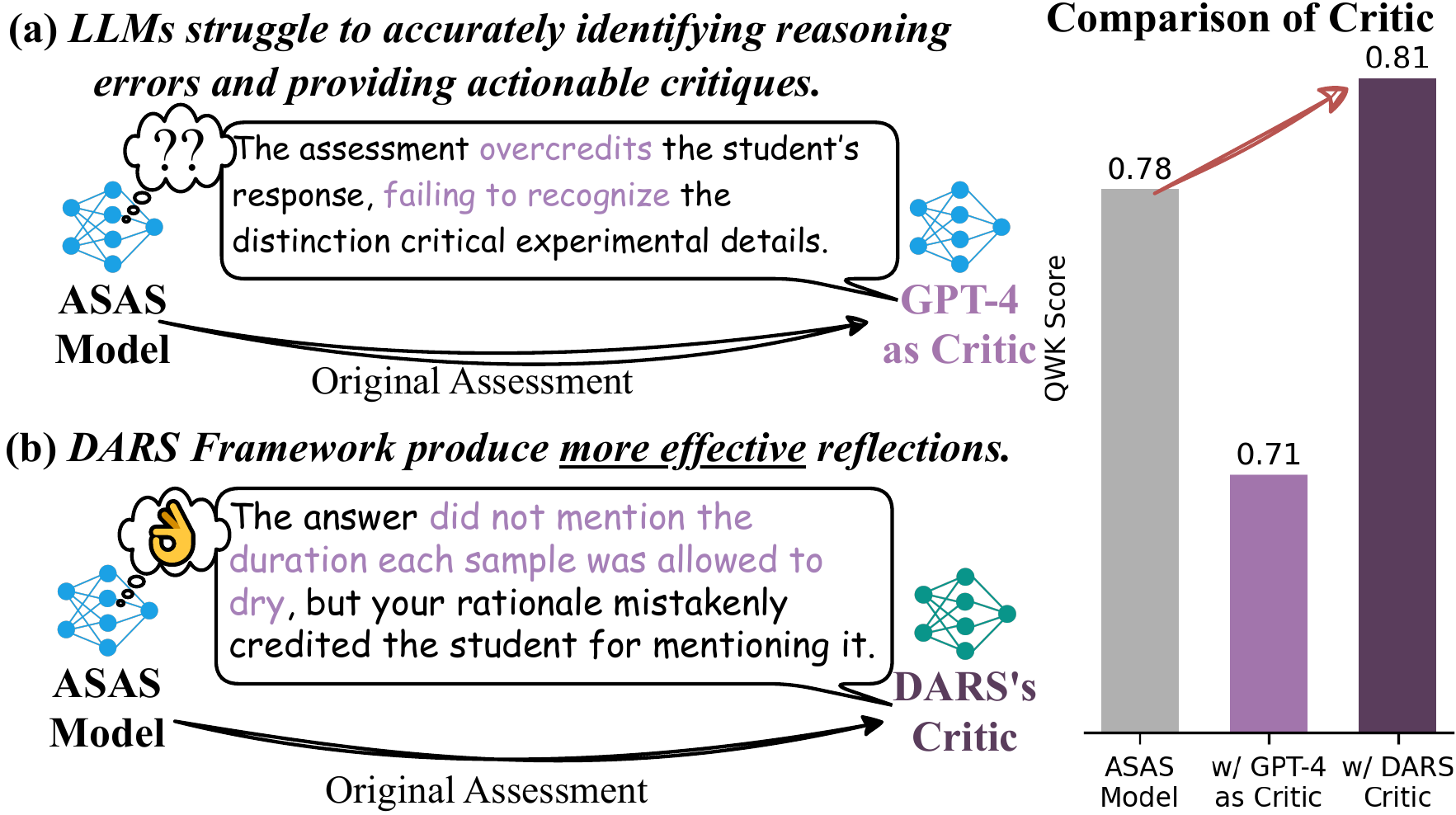}
\caption{\textbf{Left (a)}: LLMs often fail to localize reasoning errors~\citep{Huang2023LargeLM}, limiting their performance in verbal RL. \textbf{Left (b)}: \texttt{DARS} leverages a \emph{contrastive reflection synthesis} pipeline to generate precise error‑correction feedback, which guides the ASAS model to generate better scoring results with more accurate rationales. %
\textbf{Right}: While using GPT-4 as the Critic results in lower ASAS performance, our \texttt{DARS} Critic yields improved results in verbal RL.} 
\label{fig:intro}
\vspace{-4mm}
\end{figure}

Recent work has attempted to improve rationale quality by fine‑tuning LLMs with Direct Preference Optimization (DPO) on synthetic preference pairs~\cite{step_value_emnlp2024,Lu2024StepControlledDL}. While DPO captures \emph{which} assessment is preferred, it fails to explain \emph{why}~\cite{dpo_drawback, sampo, lu-etal-2025-rolemrc}, leaving key reasoning steps opaque. Verbal Reinforcement Learning (VRL) addresses the gap by explicitly critiquing and revising model reasoning~\cite{shinn2023reflexion,wei-jie-etal-2024-interpretable}. However, LLMs struggle to self-correct due to their limited ability to accurately detect and locate reasoning errors~\cite{yan-etal-2024-mirror, yan2025position}.

As illustrated in Figure~\ref{fig:intro}, evaluating whether a student answer addresses all key answer elements is non‑trivial. Even advanced models such as GPT‑4 often overlook flawed steps and produce vague, superficial reflections~\cite{kamoi-etal-2024-llms}, affecting the effectiveness of self‑correction. The lack of high-quality annotations further compounds this challenge~\cite{liu2024best}.
We argue that these limitations arise from the sequential decoding paradigm of current LLMs, which struggle to represent and reason over the graph‑like conceptual structures underlying assessment decision-making process~\cite{LeCun2022APT}. Effective self‑correction requires reasoning to be decomposed into discrete components~\cite{subramaniam2025multiagent}, akin to ``nodes'' in a graph, that can be individually inspected and revised.

To this end, we propose a \textbf{contrastive reflection synthesis} pipeline (Section~\ref{sec:method_data_curation}) that transforms preference-based reasoning path pairs into targeted, fine-grained verbal critiques without using of human annotation. Given a student response and a set of key answer elements, we construct a reasoning tree through progressive binary comparisons, where each decision reflects the presence or absence of a key answer element. By comparing the paths taken by two assessments over the \emph{same} tree, we can localize the exact nodes at which their reasoning diverges and automatically generate targeted error messages (Figure~\ref{fig:intro}, \texttt{DARS} Critic).

Building on these generated critiques, we train \texttt{DARS}, a \textbf{D}u\textbf{A}l‑model \textbf{R}eflective \textbf{S}coring framework comprising dedicated \textit{Reasoner} and \textit{Critic} models (Section~\ref{sec:reasoning_framework}). The Reasoner produces an initial score and rationale; while the Critic delivers both verbal reflection to the Reasoner and a termination token that signals convergence, enabling effective VRL without relying on oracle labels or manually-defined thresholds. %

In summary, our contributions are as follows:
\begin{enumerate}[leftmargin=*,noitemsep,topsep=0pt]
\item We propose a \emph{contrastive reflection synthesis} pipeline that \textbf{\emph{automatically transforms binary preferences into fine‑grained error‑correction reflections}}.
\item We present \texttt{DARS}, to enable \textbf{\emph{effective Verbal RL for ASAS reasoning}}. The Critic is innovatively designed to be capable of reflect reasoning errors and determining reasoning convergence.%
\item Extensive experiments show that \texttt{DARS} \textbf{\emph{consistently outperforms}} baselines, even in scarce data settings, scales with model size, and generalize across different LLM base models.
\end{enumerate}

%% file: sections/3_method.tex
\section{Preliminary}
Existing ASAS systems primarily aim to automate teachers' complex reasoning processes on the assessment of short answer questions, typically operating within a classification paradigm~\cite{grading_classification,yue-aes-2017}. Existing datasets only contain annotated student answer and score pairs. Therefore, ASAS systems take various contextual input, including \emph{question prompts}, \emph{key answer elements} (e.g., keywords or phrases that qualify for marks), \emph{marking rubrics} (e.g., criteria for assigning scores), and \emph{student responses}, and are trained to predict a \emph{score} as output.

Given a single question, the dataset can be represented as \( D = \{(x_i, y_i)\}_{i=1}^{N} \), where \( x_i \) denotes a student's response and \( y_i \) represents the corresponding score assigned by human assessors. Let \( \mathcal{K} = \{k_j\}_{j=1}^{M} \) represent the set of key answer elements for the current question, where \( M \) is the number of distinct elements expected in a complete answer. The scoring process can be formalized using a question-specific scoring function \( f_r(\cdot) \), which determines the final score based on the extend to which student's response includes the required elements:
\begin{equation}
y_i = f_r(\mathbf{v}(x_i, \mathcal{K})), \label{formula:assessment_problem_definenation}
\end{equation}
where \( \mathbf{v}(x_i, \mathcal{K}) \in \mathbb{R}^{M} \) is a multi-hot vector indicating the presence of each key element \( k_j \in \mathcal{K} \) in the student response \( x_i \).
This coverage vector is then mapped to the final score through \( f_r \). However, due to the complexity of the reasoning process and annotation costs, such intermediate assessment states are not available within current datasets.

\begin{figure*}[t]
\centering
\includegraphics[width=\linewidth]{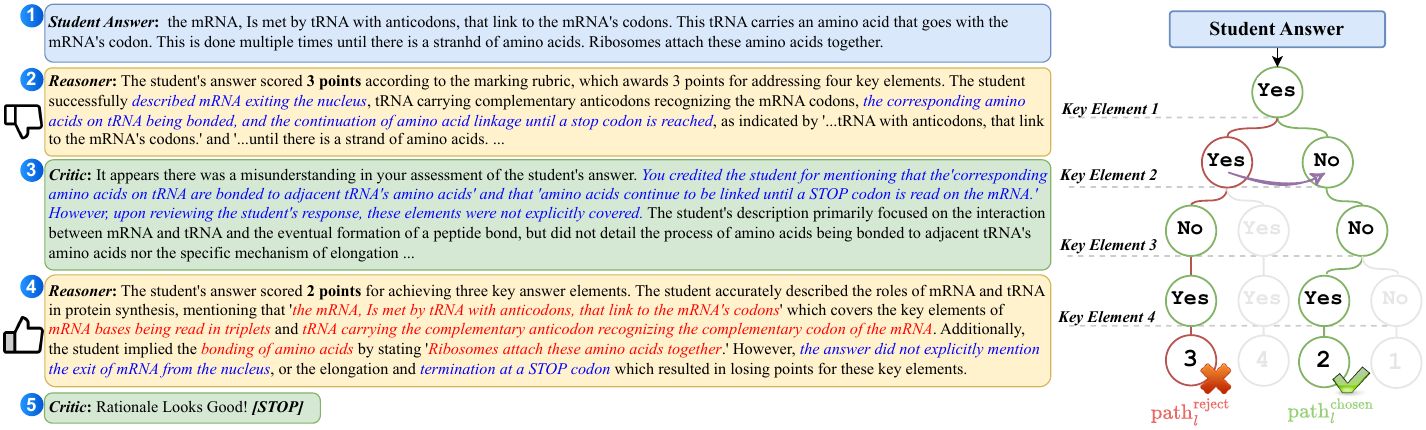}

\caption{\textbf{(Left) An example conversation between the \colorbox{rcolor}{Reasoner} and \colorbox{ccolor}{Critic} in the \texttt{DARS} framework. (Right) A thought tree constructed from a single student answer.} Structured thought tree paths are generated by an LLM and used to produce free-text reasoning outcomes (e.g., \textcircled{\raisebox{-0.3pt}{\scriptsize2}}, \textcircled{\raisebox{-0.3pt}{\scriptsize4}}). Discrepancies between distinct reasoning paths are identified and used to prompt the LLM to generate a verbal reflection (e.g., \textcircled{\raisebox{-0.3pt}{\scriptsize3}}), explicitly highlighting errors in the rejected reasoning trace. Text related to the Reasoner’s initial mistake is highlighted in \textcolor{blue}{blue}, while corrections introduced during refinement are marked in \textcolor{red}{red}. \textcircled{\raisebox{-0.3pt}{\scriptsize1}} denotes the framework’s input (question context omitted for brevity), and the final Reasoner response before Critic termination (\textcircled{\raisebox{-0.3pt}{\scriptsize4}}) represents the framework’s output. A detailed explanation of the example is provided in \textsection{\ref{sec:figure_full_example}}.}
\label{fig:main_example}

\end{figure*}

To bridge this gap in intermediate steps, a recent approach \citep{li_emnlp2024} leverages a structured thought tree generated by LLMs to mimic the human assessment process (as illustrated in Figure \ref{fig:main_example}). %
Formally, for each student answer \(x_i\) we construct an assessment decisions thought tree \(\mathcal{T}=\{\mathcal{Z}_{\ell}\}_{\ell=1}^{d}\) following \citeauthor{li_emnlp2024}  
Each distinct tree path $\mathcal{Z}_{\ell}$ encodes binary decisions over \(M\) key elements:
\begin{equation}
    \hat{\mathbf{v}}(\mathcal{Z}_{\ell}) = [z_1^{(\ell)}, z_2^{(\ell)}, \dots, z_M^{(\ell)}],
\label{formula:tree_path}
\end{equation}
where \( z_j^{(\ell)} \in \{0,1\} \) indicates whether the \( j^\text{th} \) key element is correctly answered or not. We define reasoning paths that yield a correct score as the human \emph{preferred} or \emph{chosen} path ($\mathcal{Z}_{\ell}^{\textsc{chosen}}$), and paths that yield an incorrect score as the human \emph{rejected} path ($\mathcal{Z}_{\ell}^{\textsc{reject}}$). The rationales \( r_{\textsc{chosen}} \) and \( r_{\textsc{reject}} \) are then derived by summarizing the intermediate decisions along their respective reasoning paths. %

\section{\texttt{DARS}: Dual-Model Reflective Scoring} %

We introduce \texttt{DARS}, a dual-model framework that pairs a \colorbox{rcolor}{\textbf{Reasoner} (\(\mathcal{R}\))} with a \colorbox{ccolor}{\textbf{Critic} (\(\mathcal{C}\))}. The Critic supplies explicit, free-form verbal reflections that iteratively steer the Reasoner’s thought process. The \texttt{DARS} framework adopt a two-stage design: Stage 1, \textbf{Contrastive Reflection Synthesis} (\textsection{\ref{sec:method_data_curation}}), constructs synthetic reflection data by comparing pairs of structured reasoning paths (``thought trees'') for the same student answer, to pinpoint where a rejected rationale diverges from a chosen one. Stage 2, \textbf{Dual-Model Training \& Inference} (\textsection{\ref{sec:reasoning_framework}}), uses supervised fine-tuning (SFT) to train a Reasoner and a Critic on these data. At inference, the Reasoner proposes an assessment and the Critic either provides a reflection for revision or terminates the loop.  Importantly, no tree is constructed at inference, and no reinforcement learning is used in training; the critique-and-revise behavior arises from SFT-trained models interacting on-policy at test time.

\subsection{Contrastive Reflection Synthesis}\label{sec:method_data_curation}
Human graders do not inspect an answer sequentially; instead, they mentally traverse a conceptual graph, where nodes represent key answer elements. In contrast, the sequential nature of LLM processing linearises this graph, often interleaving correct and incorrect claims, which obscures the exact source of the error. Therefore, naively prompting an LLM to reflect on its own errors typically produces vague, superficial, or uninformative rationales\footnote{We provide empirical analysis for this in \textsection{\ref{sec:main_exps}}}~\cite{Yin2024RelativePO, Jiang2024BridgingAM}.

Our pipeline restores this missing structural representation by converting each reasoning preference pair into a fine‑grained error critique that explains \emph{``why \(r_{\textsc{reject}}\) is inferior to \(r_{\textsc{chosen}}\)''} using divergent nodes to identify the \emph{minimal} sub‑graph responsible for the discrepancy. These targeted critiques give the \textit{Critic} module a precise mechanism for verbal reinforcement learning, enabling it to generate clear guidance for error correction.

According to Equation (\ref{formula:tree_path}), for each student answer \(x_i\) we construct a thought tree \(\mathcal{T}=\{\mathcal{Z}_{\ell}\}_{\ell=1}^{d}\). Nodes in $\hat{\mathbf{v}}$ inherit the partial decision vector of their ancestors, while edges represent the incremental ``reveal'' of one additional element, mirroring a breadth‑first traversal of the graph.

\subsubsection*{Step 1: Identify Discrepancy in Reasoning Paths}
Given a preference pair \((r_{\textsc{reject}},r_{\textsc{chosen}})\), we align each rationale with its original path and compute a signed \emph{difference vector}:
\[
\Delta\mathbf{v}= \hat{\mathbf{v}}\bigl(\mathcal{Z}^{\textsc{chosen}}_{\ell}\bigr)- \hat{\mathbf{v}}\bigl(\mathcal{Z}^{\textsc{reject}}_{\ell}\bigr),
\]
which captures the discrepancies between $\mathcal{Z}^{\textsc{chosen}}_{\ell}$ and $\mathcal{Z}^{\textsc{reject}}_{\ell}$. Each component $\Delta_j$ in $\Delta\mathbf{v}$ flags a node where the chosen (or \ rejected) path newly asserts the presence of the key element \(k_j\), thereby localising points of divergence.  
\[
\small
\Delta_j = 
\begin{cases} 
1 & \text{if decision for } k_j \text{ changed from 0 to 1}, \\ 
-1 & \text{if decision for } k_j \text{ changed from 1 to 0}, \\ 
0 & \text{if decision is the same}.
\end{cases}
\]
Because every \(k_j\) is tied to an explicit rubric criterion, \(\Delta\mathbf{v}\) directly identifies the sub‑graph responsible for diverging scores. We convert each non‑zero component into a natural‑language \emph{structural hint}\footnote{A detailed prompt template is provided in \textsection{\ref{box:prompt_template}}.} that highlights the differences in the intermediate assessment decisions (e.g. \(r_{\textsc{reject}}\) missed $k_j$ that the student has already included):
\begin{equation}
    \text{hint}_{\Delta\mathbf{v}} = \text{Prompt}(\Delta\mathbf{v}, \mathcal{K}).
\end{equation}

\subsubsection*{Step 2: Generate Synthetic Reflections}
After identifying discrepancies and constructing the hint prompt, we prompt an LLM (e.g., GPT-4-turbo) to generate a verbal reflection between the preference pair \( r_{\textsc{reject}} \) and \( r_{\textsc{chosen}} \):
\begin{equation}
    r_{\text{reflect}} = \texttt{LLM}_{\theta}(x_i, r_{\textsc{reject}}, r_{\textsc{chosen}}, \text{hint}_{\Delta\mathbf{v}}),
    \label{eq:r_pref}
\end{equation}
Because the hint anchors the prompt in the concept graph, the model tends to produce concise, node‑level critiques such as \textit{``You marked \textsc{Photosynthesis produces oxygen} absent, but the answer states `plants \underline{release O\textsubscript{2}},' satisfying node \(k_3\).''} We record this free‑text reflection as \(r_{\text{reflect}}\).

\begin{figure}[ht]
\centering
\includegraphics[width=0.8\linewidth]{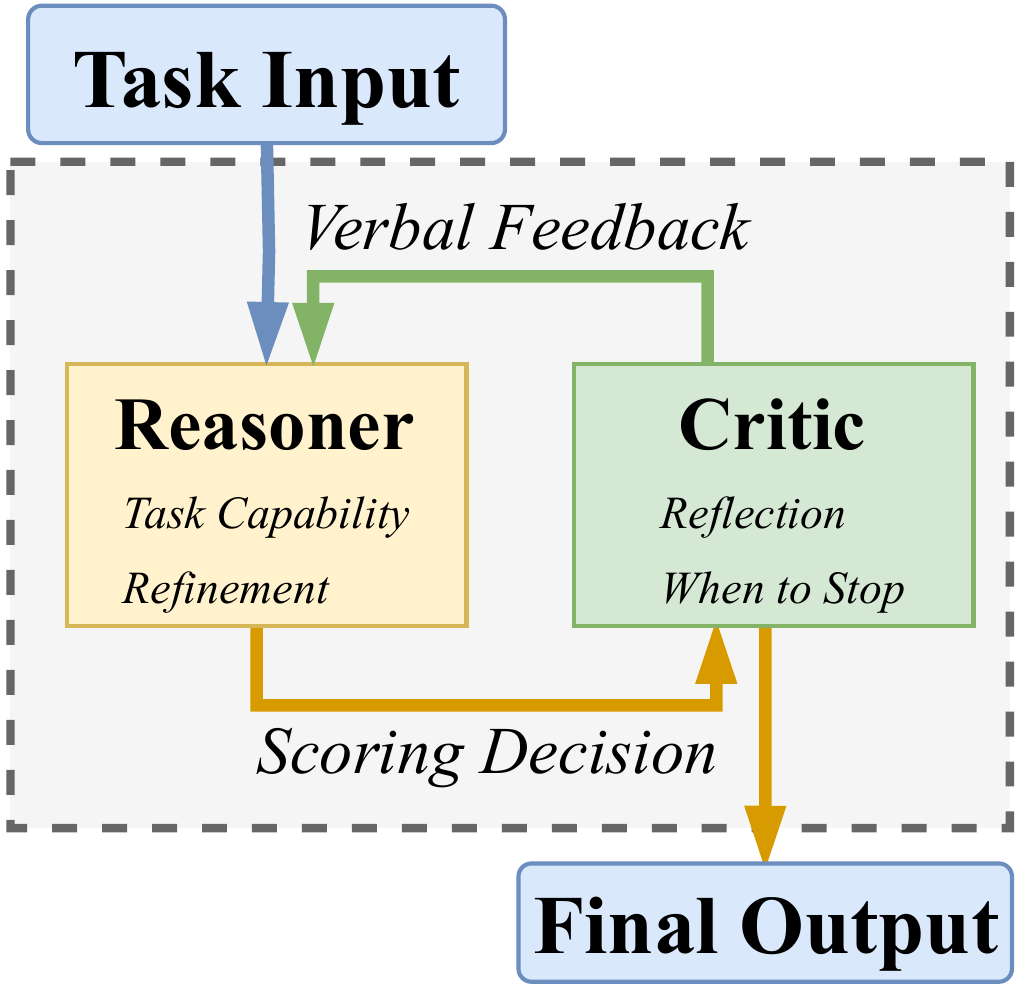} 

\caption{\textbf{Illustration of \texttt{DARS} Framework.}}
\label{fig:framework}
\end{figure}
\subsection{Dual-Model Training \& Inference} \label{sec:reasoning_framework}
Figure~\ref{fig:framework} outlines how the Reasoner and Critic cooperate at inference time. Starting from a student answer, the Reasoner drafts an initial scoring rationale. The Critic then either (i) provides a targeted reflection to prompt a revision from the Reasoner, or (ii) outputs a special \textsc{[Stop]} token to terminate the loop. This iterative dialogue continues until the Critic determines that the reasoning has converged.

\subsection*{Training Reasoner and Critic Models} \label{sec:training}
Build on the synthetic reflection data generated, we create diverse data combinations to train the Reasoner and the Critic on refinement and reflection capabilities. For clarity we reference the numbered turns in Figure~\ref{fig:main_example}.\footnote{Full implementation details are provided in \textsection{\ref{sec:further_experiment_setup}}.}

\paragraph{\colorbox{rcolor}{Reasoner (\(\mathcal{R}\))}}
The training data for the Reasoner is designed to include two capabilities:
\begin{itemize}[leftmargin=*,noitemsep,topsep=0pt,label={}]
    \item \textbf{Task Capability}: \(\mathcal{R}\) takes \textcircled{\raisebox{-0.3pt} {\scriptsize1}} (question context and student answer) as input, and predicts \textcircled{\raisebox{-0.3pt} {\scriptsize2}}  (an initial assessment $r$).
    \item \textbf{Refinement}: \(\mathcal{R}\) takes \textcircled{\raisebox{-0.3pt} {\scriptsize1}} \& \textcircled{\raisebox{-0.3pt} {\scriptsize2}} (assessment histories, e.g., $r_{\textsc{reject}}$), with \textcircled{\raisebox{-0.3pt} {\scriptsize3}} (verbal reflection generated by \(\mathcal{C}\) , e.g., \(r_{\text{reflect}}\)) as input, and predict \textcircled{\raisebox{-0.3pt} {\scriptsize4}} (an refined assessment, e.g., $r_{\textsc{chosen}}$).
\end{itemize}
\paragraph{\colorbox{ccolor}{Critic (\(\mathcal{C}\))}}
The training data for the Critic is designed to include two capabilities:
\begin{itemize}[leftmargin=*,noitemsep,topsep=0pt,label={}]
\item \textbf{Reflection}: If the assessment is incorrect, \(\mathcal{C}\) is trained to take previous assessment histories (e.g., \textcircled{\raisebox{-0.3pt} {\scriptsize1}}-\textcircled{\raisebox{-0.3pt} {\scriptsize2}} or \textcircled{\raisebox{-0.3pt} {\scriptsize1}}-\textcircled{\raisebox{-0.3pt} {\scriptsize4}}) as input, and predict \textcircled{\raisebox{-0.3pt} {\scriptsize3}} (a reflection $r_{\text{reflect}}$ for wrong assessment) as output.
\item \textbf{When to Stop}: \(\mathcal{C}\) takes \(\mathcal{R}\)'s previous assessment outcome, either from single-round \textcircled{\raisebox{-0.3pt} {\scriptsize1}}-\textcircled{\raisebox{-0.3pt} {\scriptsize2}} or multi-rounds \textcircled{\raisebox{-0.3pt} {\scriptsize1}}-\textcircled{\raisebox{-0.3pt} {\scriptsize4}} as input, and validate the correctness of the assessment. If the assessment is correct, \(\mathcal{C}\) predict \textcircled{\raisebox{-0.3pt} {\scriptsize5}}, a special token \textsc{[Stop]} that signals the termination of the reasoning loop and outputs the final assessment generated by \(\mathcal{R}\).
\end{itemize}

The Critic is trained to supply two complementary feedbacks in natural language:  
(1) \emph{Reflection} that diagnose specific reasoning flaws, and  
(2) \emph{When to Stop} that decides when the assessment has converged.  
Both capabilities are learned \emph{without} the need of oracle labels, or setting maximum iteration limits, overcoming those weaknesses in prior work~\cite{shinn2023reflexion, kim2023language}.
\subsection*{Inference-Time Iterative Refinement}
Once the Reasoner and Critic models are trained, they could collaborate to refine the assessment rationale at inference time through iterative conversations. At each iteration step \( t \), \(\mathcal{R}\) generates an assessment trajectory \(\hat{y}_r^{0}, \hat{y}_r^{1}, ...,\hat{y}_r^{T}\): 
\begin{equation}
\begin{aligned}
&\textbf{Initialization:} \quad \hat{y}_r^0 = \mathcal{R}\bigl(x_i\bigr) \\
&\textbf{Iterative~Reflection:} \quad \\
&\begin{cases}
\hat{y}_r^{(t+1)} = \mathcal{R}\Bigl(\hat{y}_r^{t},\mathcal{C}(\hat{y}_r^{t})\Bigr), & \text{if } \mathcal{C}(\hat{y}_r^{t})=\mathrm{Reflect}, \\
\hat{y}_r^T = \hat{y}_r^{t}, & \text{if } \mathcal{C}(\hat{y}_r^{t})=\textsc{[Stop]}.
\end{cases}
\end{aligned}\notag
\label{eq:reasoner_critic_update}
\end{equation}
\noindent
\(\mathcal{C}(\cdot)\) checks the correctness of \(\hat{y}_r^{t}\). If refinement is needed, it generates a verbal reflection for \(\mathcal{R}\) to refine \(\hat{y}_r^{t}\). Otherwise, \textsc{[Stop]} is triggered, and \textbf{final assessment \(\hat{y}_r^T\) from \(\mathcal{R}\)} is the output.

%% file: sections/4_experiment.tex
\section{Experiments} \label{sec:experiments}

\subsection{Experimental Setup}
\paragraph{Datasets} We use two data sources, consisting of a total of six different datasets, for our experiments: (1) The Hewlett Foundation Short Answer Scoring (ASAP) dataset~\cite{asap-aes}, which contains short essay responses across science and biology topics (we exclude essay-like or multimodal subsets); and (2) A proprietary dataset comprising student responses to biology exam questions, where human-assigned scores are provided.\footnote{Dataset statistics are in Table \ref{tab:data_statistic}.}

\begin{table*}
\centering
\resizebox{\textwidth}{!}{
\begin{tabular}{ccccccccccccccccccc}
\toprule
\multirow{2}{*}{\textbf{Methods}} & \multicolumn{3}{c}{\textbf{Classification Baseline}} & \multicolumn{9}{c}{\cellcolor{gray!20}\textbf{Generative Baselines} (\emph{Single Model Reasoning})} &  \multicolumn{6}{c}{\cellcolor{gray!20}\textbf{Dual-Model Reasoning with Critic Models}}\\ 
\cmidrule(r){2-4} \cmidrule(l){5-13} \cmidrule(l){14-19} 
& \multicolumn{3}{c}{PLM Classifier} & \multicolumn{3}{c}{SFT} & \multicolumn{3}{c}{DPO} & \multicolumn{3}{c}{(\texttt{DARS}) Reasoner only} & \multicolumn{3}{c}{GPT-4 as Critic} &  \multicolumn{3}{c}{\textbf{(\texttt{DARS}) Reasoner+Critic}}\\ 
\cmidrule(l){1-1} \cmidrule(l){2-4} \cmidrule(l){5-13} \cmidrule(l){14-19} 
\textbf{Datasets} & ACC & F1 & \textbf{QWK} & ACC & F1 & \textbf{QWK} & ACC & F1 & \textbf{QWK} & ACC & F1 & \textbf{QWK}  & ACC & F1 & \textbf{QWK} & ACC$^{\dagger,*}$ & F1$^{\dagger,*}$ & \textbf{QWK}$^{*}$\\
\cmidrule(r){1-1} \cmidrule(r){2-4} \cmidrule(l){5-7} \cmidrule(l){8-10} \cmidrule(l){11-13} \cmidrule(l){14-16} \cmidrule(l){17-19} 
\textbf{ASAP 1}    & 0.7767 & 0.7805 & 0.8528 & 0.6968 & 0.7073 & \textbf{0.8277} & 0.6895 & 0.5655 & 0.8051 & 0.6480 & 0.6606 & 0.8073 & 0.5181 & 0.5106 & 0.6349 & \textbf{0.7274} & \textbf{0.7315} & 0.8100 \\
\textbf{ASAP 2}    & 0.6798 & 0.6817 & 0.8187 & \textbf{0.7324} & \textbf{0.7468} & \textbf{0.8420} & 0.6761 & 0.6783 & 0.8033 & 0.6925 & 0.7074 & 0.8136 & 0.5869 & 0.5636 & 0.6532 & 0.7136 & 0.7303 & 0.8277 \\
\textbf{ASAP 5}    & 0.8625 & 0.6055 & 0.8187 & 0.8495 & 0.5600 & 0.8203 & 0.8612 & 0.6449 & 0.8001 & 0.8545 & 0.5424 & 0.7766 & 0.8177 & 0.5119 & 0.6340 & \textbf{0.8645} & \textbf{0.6303} & \textbf{0.8326} \\
\textbf{ASAP 6}    & 0.8891 & 0.6118 & 0.8426 & 0.8314 & 0.5513 & 0.7273 & 0.8314 & 0.5420 & 0.7522 & 0.8280 & 0.5628 & 0.7232 & 0.8130 & 0.4265 & 0.4754 & \textbf{0.8648} & \textbf{0.5988} & \textbf{0.8016} \\
\textbf{Pty 1} & 0.6787 & 0.6784 & 0.8853 & 0.5236 & 0.5197 & 0.8082 & 0.5236 & 0.4670 & 0.8196 & 0.5551 & 0.5584 & 0.8221 & 0.4134 & 0.3407 & 0.6018 & \textbf{0.5709} & \textbf{0.5653} & \textbf{0.8253} \\
\textbf{Pty 2} & 0.6224 & 0.6355 & 0.8385 & 0.5459 & 0.5377 & 0.7004 & 0.5561 & 0.5600 & 0.7599 & 0.5765 & 0.5752 & 0.7604 & 0.5357 & 0.5219 & 0.7688 & \textbf{0.6071} & \textbf{0.6059} & \textbf{0.7705} \\
\cmidrule(r){1-1} \cmidrule(r){2-4} \cmidrule(l){5-7} \cmidrule(l){8-10} \cmidrule(l){11-13} \cmidrule(l){14-16} \cmidrule(l){17-19} 
\textbf{Overall}  & 0.7515 & 0.6656 & 0.8428 & 0.6966 & 0.6038 & 0.7877 & 0.6897 & 0.5763 & 0.7900 & 0.6925 & 0.6011 & 0.7839 & 0.6141 & 0.4792& 0.6280 & \textbf{0.7247} & \textbf{0.6437} & \textbf{0.8113} \\ 
\bottomrule
\end{tabular}
}
\caption{\textbf{Comparison of assessment performance across baseline and Reasoner only preference optimization methods.} Generative methods are indicated with a \colorbox{gray!20}{gray background}. All methods were reproduced or trained using the same LLaMA 3B model as the base. We highlighted the highest values for ACC ($\uparrow$), F1 Score ($\uparrow$), and QWK ($\uparrow$) among generative methods in \textbf{bold}. The overall performance is calculated as the average across all datasets. Symbols $\dagger$ and $*$ indicate statistical significance compared to SFT and DPO by each metric, respectively.}
\label{tab:main_assessment_comparison}
\end{table*}

\paragraph{Evaluation Metrics}
We evaluate the assessment performance using Accuracy (ACC), macro F1 (F1), and Quadratic Weighted Kappa (QWK). %

\paragraph{Baselines}
We compare with four baselines:\footnote{Further details about the experimental setup are in \textsection{\ref{sec:further_experiment_setup}}.} 
\begin{itemize}[leftmargin=*,noitemsep,topsep=0pt,label={}]
\item \textbf{PLM Classifier}: A text classifier built on a pre-trained \texttt{Deberta-v3-large} model~\cite{debertav3} and fine-tuned on various datasets.
\item \textbf{SFT}: A Reasoner-only, supervised fine-tuning baseline trained with datasets released by~\citep{li_emnlp2024} (e.g, takes \textcircled{\raisebox{-0.3pt} {\scriptsize1}} as input, predicts \textcircled{\raisebox{-0.3pt} {\scriptsize2}}).
\item \textbf{DPO}: A DPO approach that performed preference optimization with synthetic reasoning preference data as presented in~\citep{li_emnlp2024} (e.g, takes \textcircled{\raisebox{-0.3pt} {\scriptsize1}} as input, optimize \textcircled{\raisebox{-0.3pt} {\scriptsize4}}$\succ$\textcircled{\raisebox{-0.3pt} {\scriptsize2}}). The base model used is the SFT baseline. 
\item \textbf{GPT-4 as Critic} A dual-model %
VRL baseline~\cite{dong-etal-2024-pace}, where Reasoner is trained within our framework, and \texttt{gpt-4-turbo} is used as the Critic to give verbal reflection instructions (e.g,  \textcircled{\raisebox{-0.3pt} {\scriptsize3}}\&\textcircled{\raisebox{-0.3pt} {\scriptsize5}} are generated by GPT-4).
\end{itemize}

\subsection{Overall Comparison} \label{sec:main_exps}
In this section, we provide a comprehensive evaluation of both scoring performance and rationale quality. As shown in Table \ref{tab:main_assessment_comparison}, we compare our dual-model reasoning framework (\texttt{DARS}) against four baselines, including both classification and generative approaches. All methods, including ours, were trained using the same LLaMA 3B model. Our results indicate that %
\emph{our framework overcomes the data scarcity issue, maintains balanced improvements across all evaluation metrics and outperforms state-of-the-art Reasoner-only and preference optimization methods}. Furthermore, our Critic model proves to be more effective than the `GPT-4 as Critic' baseline, highlighting its ability to provide more specialized and precise reflection to guide the Reasoner model.

\paragraph{Classifier Baseline} The PLM Classifier serves as a strong baseline as it is directly fine-tuned on student answer scoring data. While it exhibits strong performance across all metrics, the \emph{classification approach lacks explainability}, as it only generates scores without providing rationales.

\paragraph{Single Model Reasoning Baselines} The Reasoner-only baselines, including SFT and DPO, aim to improve explainability by generating rationales for scoring decisions. However, these methods generally underperform compared to classification-based approaches, particularly on the proprietary datasets, where \emph{data scarcity presents a major challenge}. The preference optimization method consistently shows modest improvements over the SFT base model in terms of QWK scores. However, \emph{these improvements come at the cost of declines in F1 (-4\%) and ACC scores (-1\%)}, suggesting a tendency to overfit to preference annotations~\cite{chowdhury2024provably,mitchell2023note}. Moreover, the \emph{implicit preference optimization process lacks transparency}, making the Reasoner-only DPO approach less reliable.

\paragraph{GPT-4 as Critic Baseline} We also evaluate a dual-model variant where GPT-4 serves as the Critic to generate reflection-based instructions for refinement. However, after multiple refinements, performance significantly declined across all datasets and evaluation metrics (\texttt{DARS} Reasoner only vs. GPT-4 as Critic). This indicates that despite GPT-4’s strong general capabilities, \emph{it struggles to produce specialized and precise reflections for refining the Reasoner's output}\footnote{Detailed case studies are provided in Appendix \ref{sec:gpt4o_case_study}.}.

\paragraph{Ours \texttt{DARS} Framework} \texttt{DARS} \emph{demonstrates significant improvements from the initial to the final iteration across all datasets, highlighting the efficacy of dual model reasoning, and test-time rationale refinement}. The \texttt{DARS} Reasoner only performance is measured on the Reasoner's \emph{first-pass predictions} (e.g. Reasoner predicts \textcircled{\raisebox{-0.3pt} {\scriptsize2}} based on \textcircled{\raisebox{-0.3pt} {\scriptsize1}}), while the Reflect w/ Critic results are generated from \texttt{DARS}, i.e. %
the final refined Reasoner output before the loop is terminated by the Critic model (e.g. \textcircled{\raisebox{-0.3pt} {\scriptsize4}}). Compared to the preference optimization baseline (SFT to DPO), our framework ((\texttt{DARS}) Reasoner only to Reasoner+Critic) not only \textbf{\emph{outperforms on average ACC, F1, and QWK scores}} but also \textbf{\emph{maintains a balanced enhancement across all metrics even under data scarcity}} (improved 5\% for ACC, 11\% for F1, and 2\% for QWK). Compared with GPT-4 as the Critic, our Critic model more effectively reflects on wrongly assessed rationales and guides the Reasoner outputs to be closer to the oracle labels (18\%-34\% better in metrics). Specifically, Reasoner+Critic surpasses the Reasoner only assessment result across all datasets and metrics (3\%-7\% improvement). Statistically, Reasoner+Critic significantly outperforms the state-of-the-art baselines (SFT and DPO)\footnote{A one-tailed t-test yielded a \emph{p}-value of $\leq 0.05$, indicating statistical significance.}.

 \begin{figure}[ht]
\centering
\includegraphics[width=\linewidth]{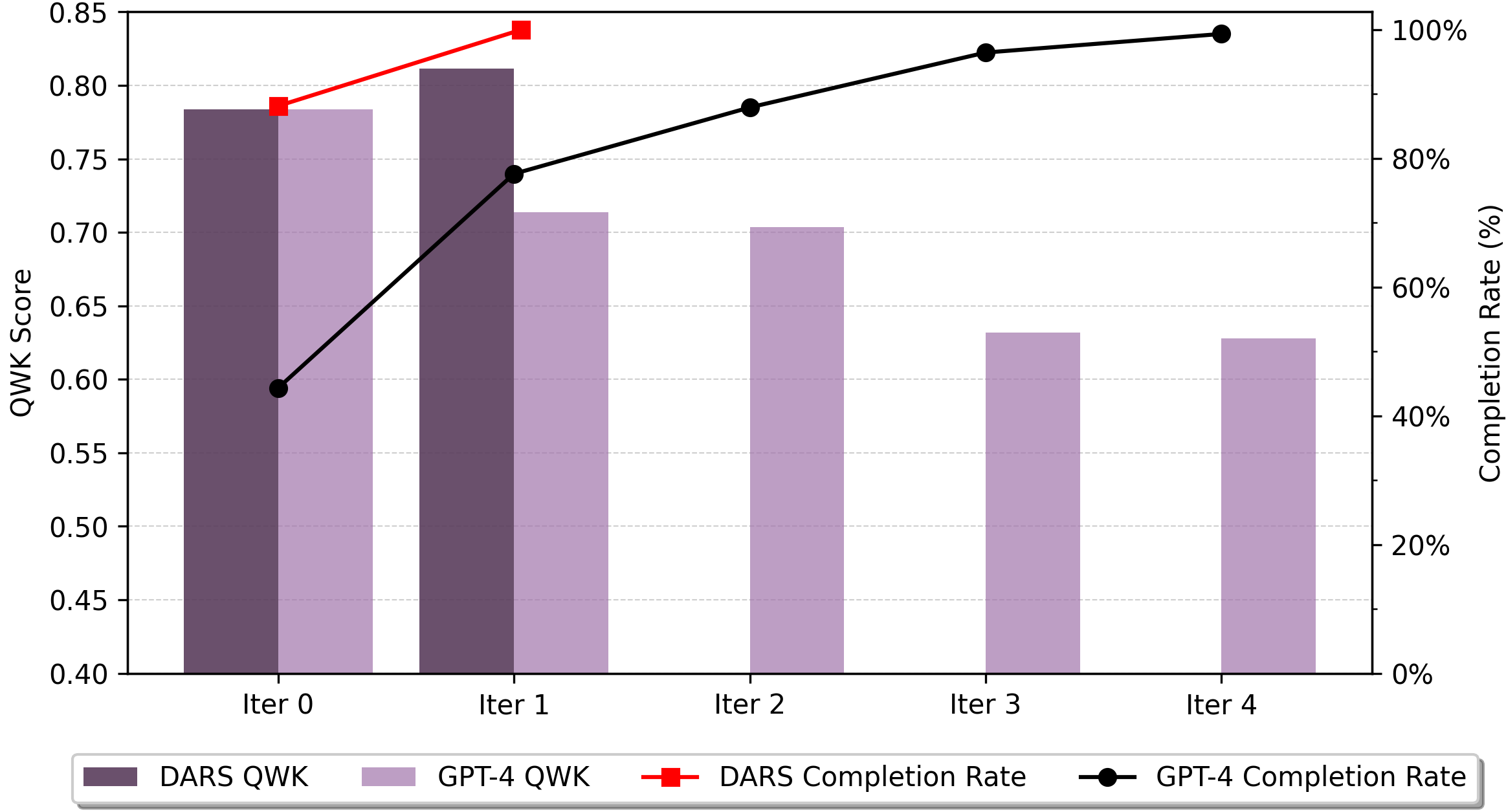}
\caption{\textbf{Performance and completion rate, where DARS outperforms GPT-4 with less iterations. }}

\label{fig:iteration}
\end{figure}

To show the effectiveness of our Critic model in reflection and determine when to stop, as illustrated in Figure \ref{fig:iteration}, we visualize the performance trend and completion rate comparison between \texttt{DARS}’s iterative reasoning process and GPT-4 as the Critic model. Our method requires only two iterations to achieve a significant improvement over iteration 0-the Reasoner’s initial prediction. In contrast, GPT-4 takes nearly four iterations to reach termination, and shows a clear trend of performance degradation across all metrics as the iterations progress.

\vspace{-2mm}
\subsection{Quality Evaluation for Reflection}
\begin{figure}[ht]
\centering
\includegraphics[width=\linewidth]{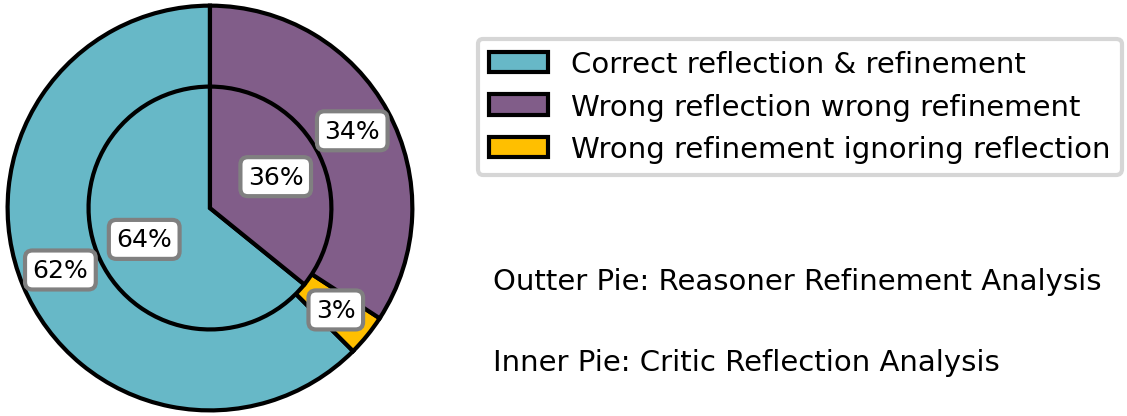}
\caption{\textbf{Qualitative analysis on reflection and refinement.}}
\label{fig:human_evaluation}
\end{figure}

To further analyze the transparency and correctness of the generated reflections, we conducted a human evaluation of the Critic-Reasoner interactions. We assessed the quality of the Critic's reflections and the subsequent Reasoner's refinements. The evaluation results are visualized in Figure \ref{fig:human_evaluation}. %

Our findings indicate that the Critic model accurately identified assessment errors in 64\% of cases, effectively localizing errors in scoring rationales. This aligns with previous observations~\cite{tyen-etal-2024-llms}, which suggest that LLMs can correct errors when provided with proper error localization. However, in 36\% of cases, the Critic's reflections were inaccurate, often due to misinterpretation of the student's answer and the scope of the key answer elements. Such inaccuracies had cascading effects: in 34\% of cases, the Critic's incorrect guidance misled the Reasoner, leading to further wrong assessments. We also observed that in 3\% of instances, the Reasoner ignored the Critic's feedback (despite correct or incorrect) %
and still produced erroneous outcomes.%
These results indicate that our Reasoner can follow the Critic's guidance 97\% of the time for refinement. Overall, \emph{these results highlight the critical role of a strong Critic for generating explainable, verbal reflection instructions, so that the Reasoner could effectively refine its predictions}. Further error analysis (\textsection{\ref{sec:our_detail_error_analysis}}) and case studies (\textsection{\ref{sec:our_case_study}}) are provided in the
Appendix.

\subsection{Scaling Experiment for \texttt{DARS} Framework} \label{sec:scaling_law}
Given that our Reasoner and Critic models are trained independently, we study the effect of model size on the performance of \texttt{DARS} using four Qwen model variants (3B, 7B, 14B, and 32B)~\cite{qwen2.5}. We trained each model using identical datasets, training procedures, and hyper-parameters, resulting in a total of 16 distinct Reasoner and Critic combinations.
\begin{figure}[!ht]
\centering
\includegraphics[width=\linewidth]{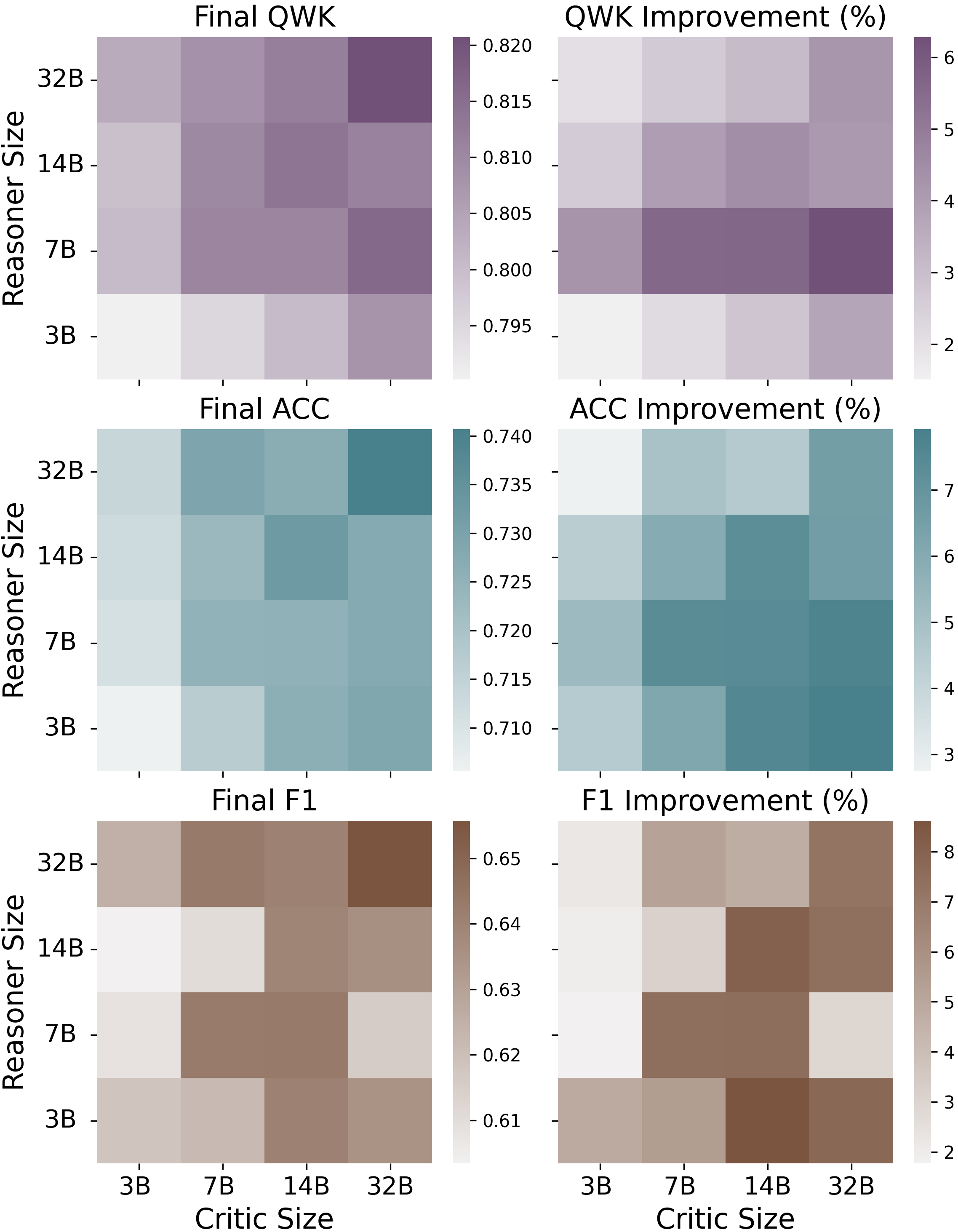} %
\caption{\textbf{Scaling experiments for \texttt{DARS}.}}
\label{fig:scaling_law}
\end{figure}

We present the overall performance and performance improvements\footnote{Performance improvement is expressed as a percentage increment compared to the Reasoner only's performance.} in Figure \ref{fig:scaling_law}. Unlike observations in prior studies~\citep{welleck2023generating, akyurek-etal-2023-rl4f, paul-etal-2024-refiner}, \textbf{\emph{our findings suggest that increasing the Critic's size}} (horizontal direction, left to right) \textbf{\emph{leads to greater performance gains (ACC and QWK), more so than increasing the Reasoner's size}} (vertical direction, bottom to top). This suggests that a larger Critic provides more precise evaluation and reflection, which the Reasoner relies upon for refinement\footnote{See \textsection{\ref{sec:compare_critic_sizes}} for case studies.}. Although larger Critic models generally improve F1 scores, this trend is not as pronounced, due to imbalances in dataset sizes and label distributions\footnote{Significant label imbalances in some datasets may cause the Reasoner to modify initially ``correct'' minority label categories, thereby affecting the overall F1 trend.}.

\subsection{Ablation Studies on \texttt{DARS}}
\begin{figure}[!h]
\centering
\includegraphics[width=\linewidth]{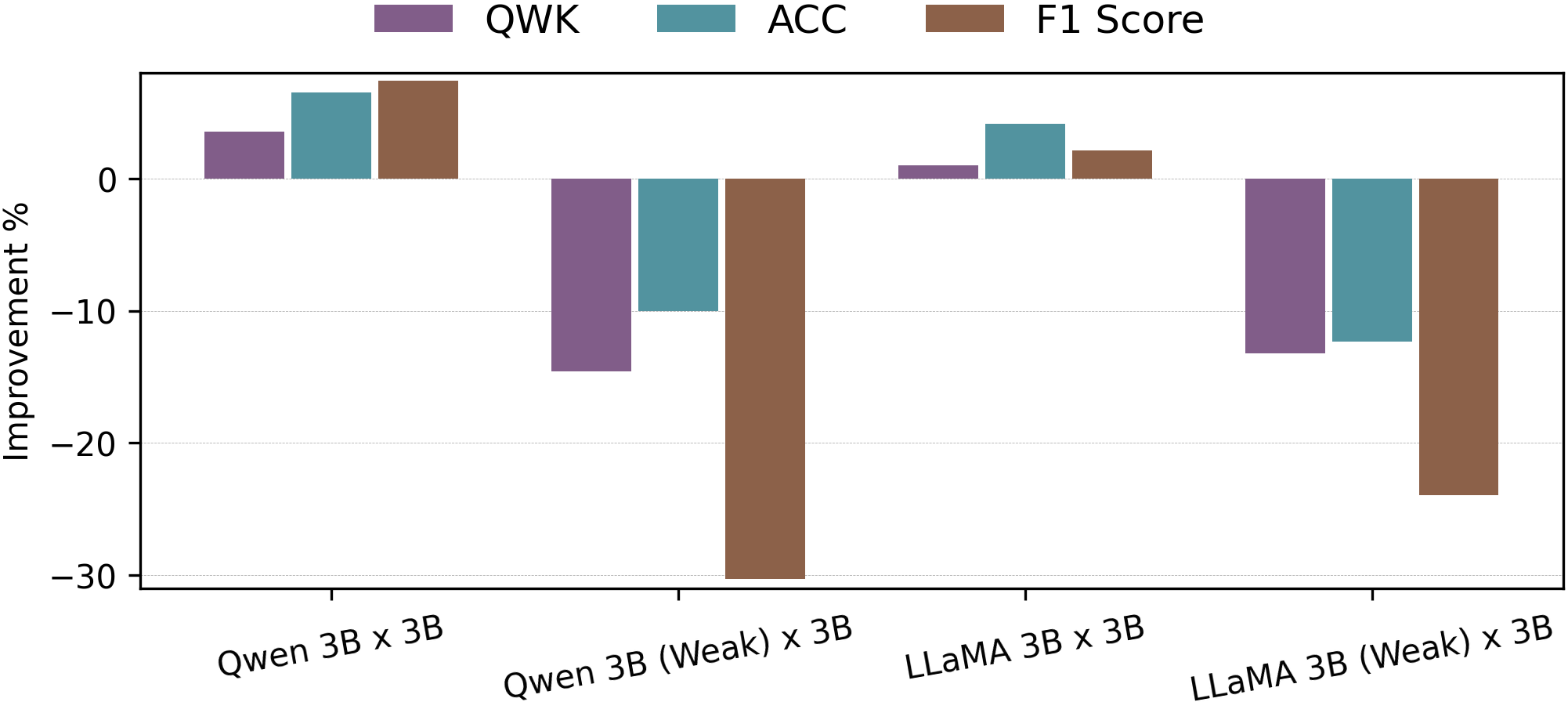}
\caption{\textbf{\texttt{DARS} refine with ``weak'' Reasoner model.}}
\label{fig:weak_reasoner}
\end{figure}

\paragraph{Can the Reasoner Refine Effectively Without Strong Task Capability?}
To investigate whether the Reasoner can perform refinement without a strong task capability, we trained two ``weak'' Reasoners with Qwen 3B and LLaMA 3B with weaker rationale training data\footnote{We characterized the data as weaker data for two reasons: (1) the rationales were sourced from ChatGPT, whereas the current training data was curated using GPT-4; (2) a previous study \cite{li_emnlp2024} shows models trained on this dataset exhibit significantly low and imbalance performance.}, following \citet{li_emnlp2023}.
As shown in Figure \ref{fig:weak_reasoner}, all the \texttt{DARS} frameworks with a ``weak'' Reasoner dropped more than 10\% in overall performance across all metrics, even with access to high-quality reflection data and a strong Critic model. %
This result shows that \textbf{\emph{without a strong task capability, the Reasoner cannot perform refinement effectively}}.

\begin{figure}[!h]
\centering

\includegraphics[width=\linewidth]{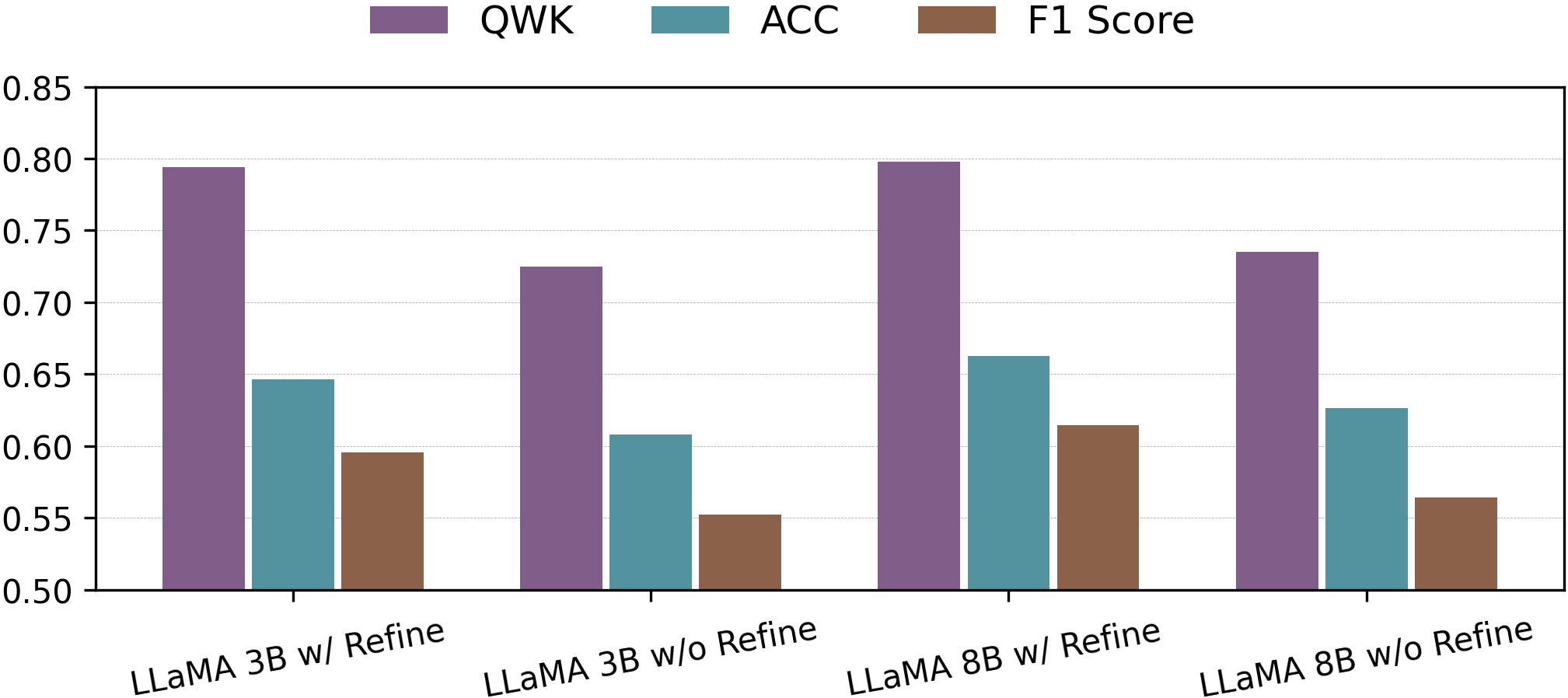}

\caption{\textbf{Ablation on the refinement data for Reasoner.}}
\label{fig:refine_unseen}
\end{figure}

\paragraph{Does Refinement Ability Benefit Reasoner's Task Capability?}
To further investigate the impact of refinement data on task performance, we trained two models: LLaMA 3B w/o Refinement and LLaMA 8B w/o Refinement by excluding the multi-turn reflection refinement data from the Reasoner's training sets.
We report the Reasoner-only's performance in Figure \ref{fig:refine_unseen}.
We observe that evaluation result for Reasoner's w/o refinement models dropped nearly 5\% in all metrics compared with including refinement data, \textbf{\emph{indicating the error correction data (e.g. training the model to refine from errors) can boost the Reasoner’s task capability}}.
This observation align closely with previous findings~\cite{tong-etal-2024-llms,kamoi-etal-2024-llms}.
We also show that reflection data can effectively regulate preference optimization training in \textsection{\ref{sec:rationale_dpo}}.

\begin{figure}[!h]
\centering
\includegraphics[width=\linewidth]{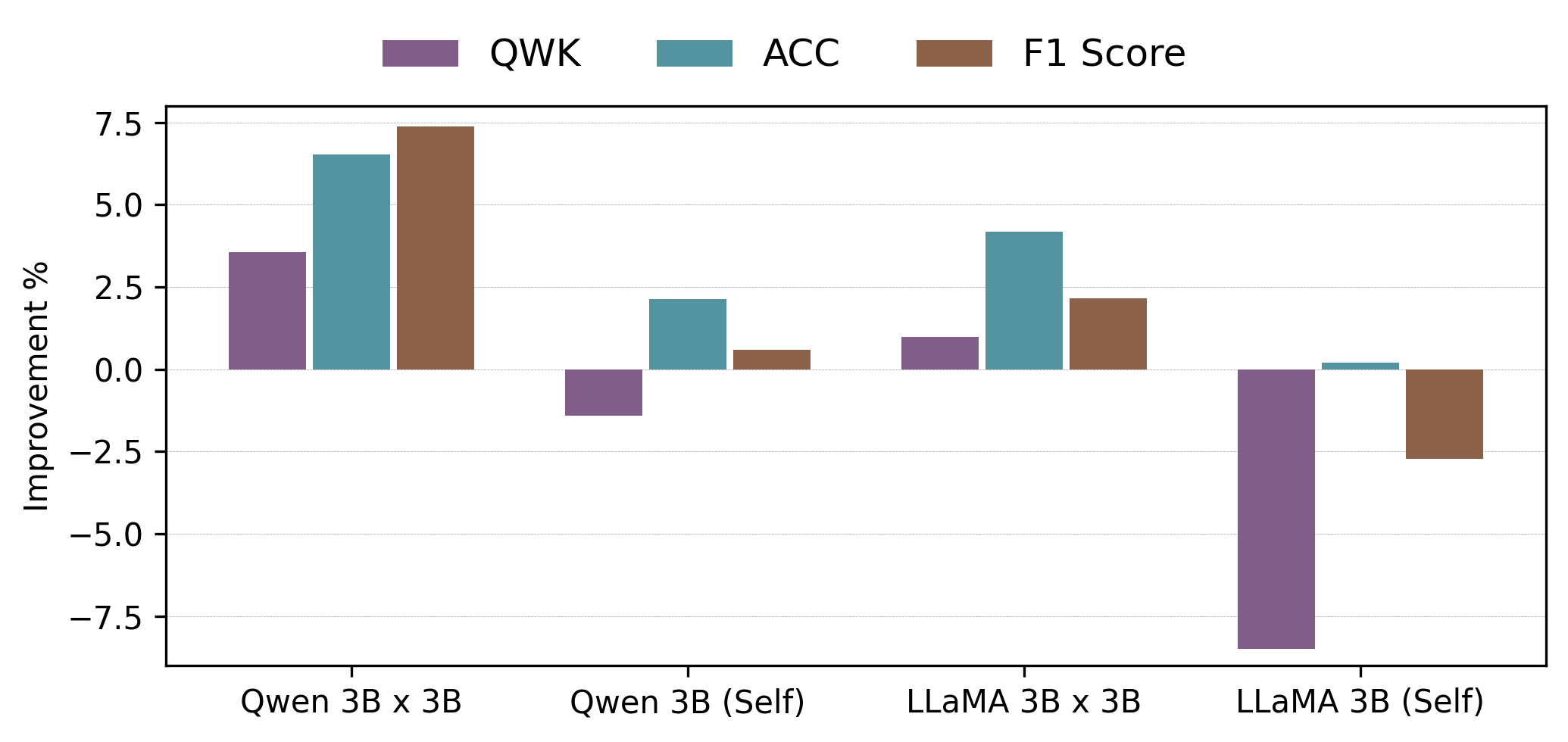}
\caption{\textbf{Combine dual-model into a single one.}}
\label{fig:self_reflection}
\end{figure}

\paragraph{Can a Single Model Perform Both Reasoning and Reflection?}
We explore whether merging the training data of both the Reasoner and Critic to train a single model would enable effective self-reflection. We trained two self-reflection models Qwen 3B (Self) and LLaMA 3B (Self). Figure \ref{fig:self_reflection} shows a significant decline in the iterative refinement process, with a negative performance improvement rate. This unified model struggles to accurately determine when to terminate the refinement process and failed to provide useful reflection instructions. These findings align with prior observations~\cite{Huang2023LargeLM}, suggesting that \textbf{\emph{``two heads are better than one''--a single model cannot effectively balance both reasoning and critique}}. %

\subsection{Generalization Studies}

\begin{figure}[!h]
\centering

\includegraphics[width=\linewidth]{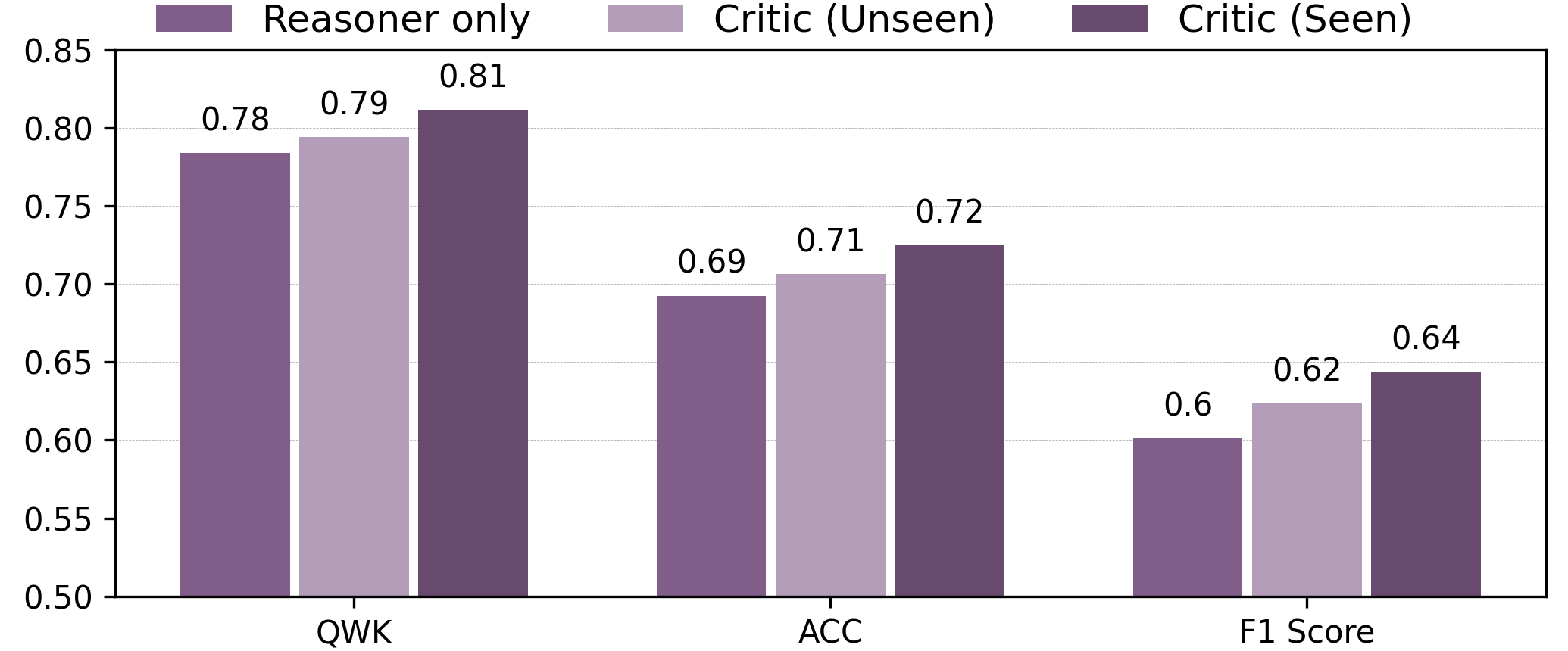}

\caption{\textbf{\texttt{DARS} Critic reflects on unseen questions.}}
\label{fig:unseen_question}
\end{figure}

\paragraph{Can Critic Effectively Reflect on Unseen Questions?} In Figure \ref{fig:unseen_question}, we evaluate the ability of the Critic model to generalize to unseen questions. To do this, we %
trained two versions of Critic: one with exposure to our proprietary datasets (Critic Seen) and one without (Critic Unseen). We use LLaMA 3B as the base model. Our results reveal that the Critic Unseen model, \textbf{\emph{despite its lack of exposure to all datasets, still enhances the Reasoner’s original assessments}} (+1\% in QWK), albeit with slightly reduced effectiveness compared to the Critic Seen model (-3\% in QWK). These findings show that %
the Critic can still provide meaningful feedback even when it has not been explicitly trained on new data.%

\begin{figure}[!h]
\centering
\includegraphics[width=\linewidth]{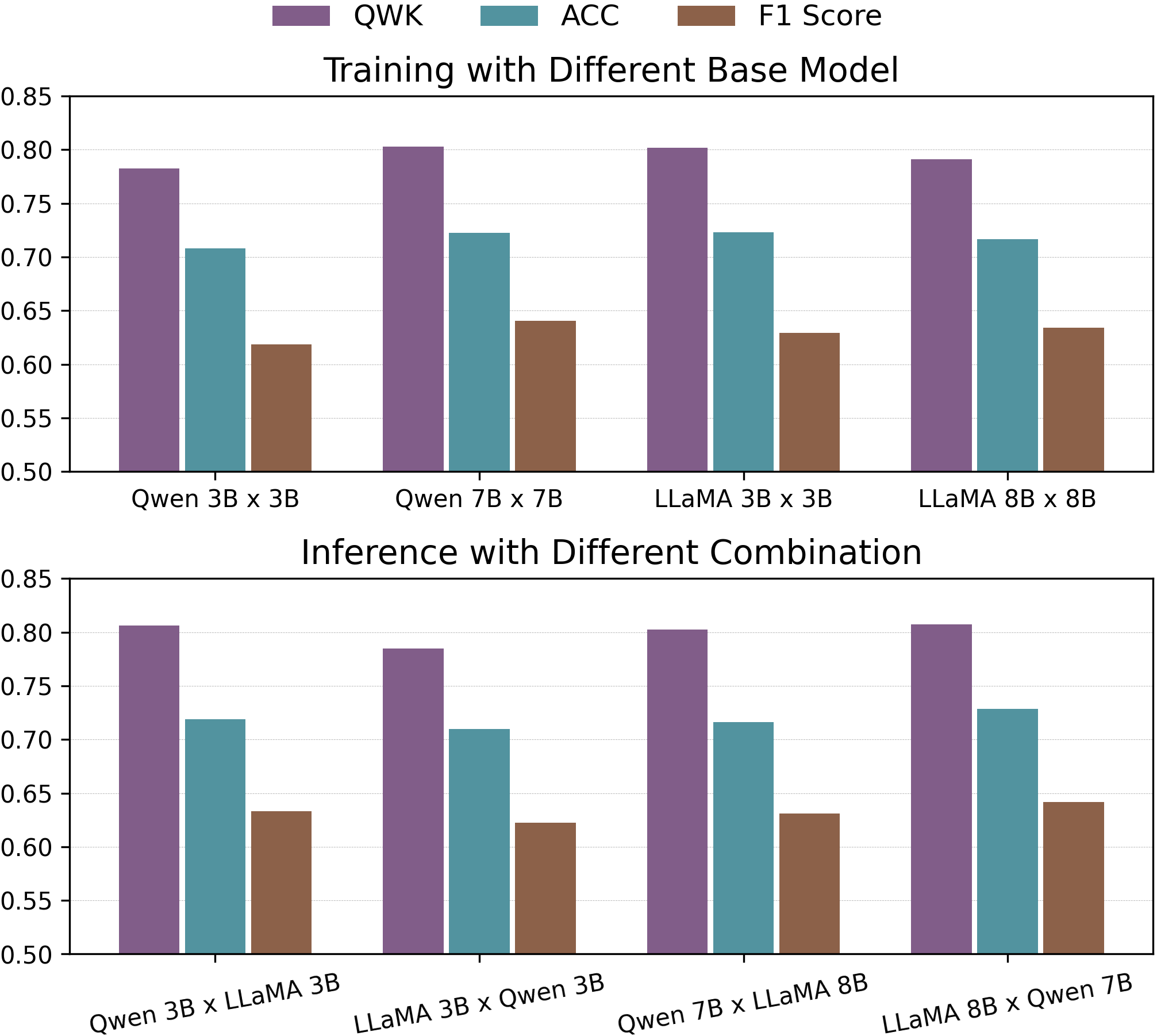}
\caption{\textbf{Generalization analysis on size, architecture and inference combinations.}}
\label{fig:generalization}
\end{figure}

\paragraph{Adaptability Beyond Model Sizes and Architectures} 
Figure \ref{fig:generalization}(a) illustrates our exploration of the performance across various base models, including LLaMA 3B, 8B and Qwen 3B, 7B. The results show minimal variance in performance across different model sizes and architectures, demonstrating that our \textbf{\emph{training method is highly adaptable}}. %
Furthermore, Figure \ref{fig:generalization}(b)  explores the feasibility of using different base models for the Reasoner and Critic at inference time, such as pairing a Qwen Reasoner with a LLaMA Critic. Our findings indicate  \textbf{\emph{consistent performance irrespective of model combinations}}. This highlights the robustness of our framework, due to its \emph{use of text} for effective interactions between Critic and Reasoner.

%% file: sections/2_related_work.tex
\section{Related Work}

\paragraph{Verbal Reinforcement Learning for Self-Reflection}
VRL has emerged as a promising approach for enhancing LLM reasoning at inference time~\cite{Huang2023LargeLM, kamoi-etal-2024-llms}. Early methods relied on self-reflection mechanisms where LLMs refined outputs using contextual cues~\cite{chen2024teaching, jiang-etal-2023-active, welleck2023generating}. However, studies show that LLMs struggle to self-correct reliably~\cite{li-etal-2024-hindsight, tyen-etal-2024-llms, chen2024can, kamoi2024evaluating}. To address this, trained critic models have been used to generate verbal feedback for LLM correction~\cite{welleck2023generating, akyurek-etal-2023-rl4f, paul-etal-2024-refiner}, though they primarily focus on single-step feedback. More complex reasoning tasks typically rely on Oracle labels for correction~\cite{shinn2023reflexion, kim2023language}. Our work introduces a dual-model framework where a Critic independently provides more detailed, trace-level reflections, eliminating the need for Oracle labels in verification.

\paragraph{Explainable Automated Student Answer Scoring}
ASAS is traditionally treated as a text classification problem~\cite{grading_classification, taghipour-ng-2016-neural}, with efforts to improve transparency via feature analysis~\cite{dong-zhang-2016-automatic, bert_feature,li-uncertainty-interpretation} and attention visualization~\cite{helen-aes-2016, yang-etal-2020-enhancing}. Recent approaches incorporate rationale generation for enhanced explainability and transparency~\cite{li_emnlp2023, learnlens} but often underperform compared to classification-based methods. \citet{li_emnlp2024} proposed a thought tree framework to model human assessment processes, leveraging LLMs for structured scoring rationales. Our work builds upon this by not only explaining decisions but also improving the transparency of assessment refinement process, through iterative LLM reasoning improvements.

%% file: sections/5_experiment_setup.tex
\section{Further Experiment Setup} \label{sec:further_experiment_setup}

This section provides additional details on the setup of the experiment:

\paragraph{Dataset Statistic}
We provide the detailed dataset statistics in Table \ref{tab:data_statistic}.
\begin{table}[h]
\centering
\resizebox{\linewidth}{!}{
\begin{tabular}{lcccc}
\toprule
\textbf{Datasets (Subjects)} & Train & Validation & Test & Score Range\\
\midrule
ASAP 1 (Science)    & 1,337 & 331  & 554 & 0-3 \\
ASAP 2 (Science)    & 1,018 & 252  & 426 & 0-3 \\
ASAP 5 (Biology)    & 1,436 & 359  & 598 & 0-3 \\
ASAP 6 (Biology)    & 1,437 & 359  & 599 & 0-3 \\ 
Proprietary 1 (Biology) & 440   & 89   & 254 & 0-4 \\
Proprietary 2 (Biology) & 358   & 72   & 196 & 0-3 \\
\bottomrule
\end{tabular}}
\caption{\small \textbf{Dataset statistics.}}
\label{tab:data_statistic}
\end{table}

\paragraph{Proprietary Dataset} The dataset provided by our project partner, a reputable national examination service. They applied a strict anonymization process before sharing the data with us. While we can report our experimental results using this data without share it with others.

\paragraph{Classification Baseline} The input to the text classifier consists of concatenated question-related information (including the question prompt, key answer elements, and marking rubric) along with the student answer, separated by newlines. The classifier is trained to predict scores. Following previous studies, we trained a separate model for each dataset and evaluated it using the original test splits~\cite{bert_classifer_aes}. We employed DeBERTa-v3-large as the base pre-trained language model~\cite{debertav3}. The reported results are averaged over five runs with different random seeds (210, 102, 231, 314, 146). The hyper-parameter settings are provided in Table~\ref{tab:hyperparameters_classification}.

\begin{table}[h]
\small
\centering
\begin{tabular}{|l|c|}
\hline
\textbf{Hyperparameter} & \textbf{Value} \\
\hline
Learning Rate          & 2e-5          \\
Batch Size             & 16            \\
Epochs                 & 15            \\
Warmup Steps           & 100           \\
Weight Decay           & 0.1           \\
Optimizer              & Adam          \\
Adam Epsilon           & 1e-8          \\
\hline
\end{tabular}
\caption{\small\textbf{Classification hyper-parameters setting.}}
\label{tab:hyperparameters_classification}
\end{table}

\paragraph{Generative Baselines} For generative baselines, the input to the model comprises the question context and student answers, with the model generating assessment rationales in textual form. The results are averaged over three runs with different random seeds. Unlike prior work~\cite{li_emnlp2024}, we conducted full parameter training using bfloat16 precision. All generative models were trained using the LLaMA-factory framework~\cite{zheng-etal-2024-llamafactory}. The hyper-parameter settings are provided in Table~\ref{tab:hyperparameters_generative}.

\begin{table}[h]
\small
\centering
\begin{tabular}{|l|c|c|}
\hline
\textbf{Hyperparameter} & \textbf{SFT} & \textbf{DPO} \\
\hline
Learning Rate          & 1e-5     & 1e-5        \\
Batch Size             & 4        & 4           \\
Gradient Accumulation  & 4        & 4           \\
Epochs                 & 4.0      & 3.0         \\
Warmup Ratio           & 0.1      & 0.1         \\
LR Scheduler Type      & cosine   & cosine     \\
Optimizer              & Adam     & Adam       \\
Adam Epsilon           & 1e-8     & 1e-8        \\
DPO ftx                & -        & 0.5         \\
DPO $\beta$            & -        & 0.1         \\
\hline
\end{tabular}
\caption{\small\textbf{Generative hyper-parameters setting.}}
\label{tab:hyperparameters_generative}
\end{table}

\paragraph{API Use for Synthetic Data Generation} We utilized \texttt{gpt-4-turbo}~\cite{gpt4} as the LLM to generate synthetic reflection data, as described in \textsection{\ref{sec:method_data_curation}}. All inference parameters were kept at their default values. The prompt template is presented in Figure \ref{box:prompt_template}~\cite{li2023overprompt}.

\begin{figure}
\begin{tcolorbox}[
    colback=gray!10,      %
    colframe=gray!80,     %
    title=Template Prompt for Generate Reflection,
    fonttitle=\bfseries,  %
    rounded corners,
    boxrule=0.5mm,        %
    width=\linewidth
]
\scriptsize
Here is an incorrect assessment rationale for the student answer:\\
\text{[Student Answer]}:\{student\_answer\}\\
Incorrect Rationale: \{reject\_rationale\}\\
This wrong rationale missed the following key elements:\\
- \{idx\}: The student didn't answer the "{key\_element[idx]}" but the incorrect rationale wrongly assessed the student mentioned it.\\
- \{idx\}: The student answered the "{key\_element[idx]}" but the incorrect rationale wrongly assessed the student didn't mention it.\\
Please construct a **reflection guidance** that\\
1. point out the incorrectly assessed key elements,\\
2. guide the model to reflect on the mistakes for generating a better assessment rationale,\\
3. pretend you are talking with an assessor using pronouns like "you",\\
4. By the end of the guidance ask the model to reflect or revise based on the feedback and retry or regenerate the rationale.\\
Output the guidance in JSON format:\{ "guidance": "..." \}
\end{tcolorbox}
\caption{\textbf{The Prompt Template for Contrastive Reflection Synthesis.}}
\label{box:prompt_template}
\end{figure}

\paragraph{\texttt{DARS} Framework} We trained both the Reasoner and Critic models using full parameters training with bfloat16 precision. All models were evaluated using greedy decoding. Except for the scaling experiment, all results were averaged over three different runs. The hyper-parameter settings are provided in Table~\ref{tab:hyperparameters_ours}. We train the Reasoner and Critic models using synthetic data we generated, as introduced in our methodology part. All those models are solely trained on the original train split, as shown in Table \ref{tab:data_statistic}. The validation split was only used to select the best checkpoint, and the Test split was never seen by the model until the evaluation.

\begin{table}[h]
\small
\centering
\begin{tabular}{|l|c|}
\hline
\textbf{Hyperparameter} & \textbf{Value} \\
\hline
Learning Rate          & 2e-5      \\
Batch Size             &           \\
- Model Size $\leq$ 8B & 16        \\
- Model Size $>$ 8B    & 8         \\
Gradient Accumulation  &           \\
- Model Size $\leq$ 8B & 1         \\
- Model Size $>$ 8B    & 2         \\
Epochs                 & 1.0       \\
Warmup Ratio           & 0.05      \\
Weight Decay           & 0.02      \\
LR Scheduler Type      & cosine    \\
Optimizer              & Adam      \\
Adam Epsilon           & 1e-8      \\
\hline
\end{tabular}
\caption{\small\textbf{\texttt{DARS} framework hyper-parameters settings.}}
\label{tab:hyperparameters_ours}
\end{table}

\paragraph{API Use for GPT-4-turbo Critic Baseline} We utilized \texttt{gpt-4-turbo-2024-04-09}~\cite{gpt4} as the Critic LLM to generate reflection data. The temperature is set as 0.7 and the maximum token generation is limited to 1,024. The prompt template is presented in Figure \ref{box:critic_template}.

\begin{figure}
\begin{tcolorbox}[
    colback=gray!10,      %
    colframe=gray!80,     %
    title=Prompt Template for GPT-4-turbo,
    fonttitle=\bfseries,  %
    rounded corners,
    boxrule=0.5mm,        %
    width=\linewidth,
    label={box:prompt}
]
\scriptsize
Given the provided assessment of the student's answer, generate constructive and actionable feedback to help the assessment model improve their response. The feedback should:\\
	1.	Highlight Areas for Improvement: Point out specific aspects where the model can enhance their assessment, such as accuracy, completeness, clarity, or structure.\\
	2.	Provide Actionable Suggestions: Offer clear, practical steps the model can take to address identified weaknesses and improve their understanding.\\
Please generate feedback based on these guidelines to guide the model in refining their response effectively.\\
If the assessment seems good enough, please output ``\text{[STOP]}'' to indicate the end of the feedback.
\end{tcolorbox}
\caption{\small\textbf{Prompt template for GPT-4-turbo as critic.}}
\label{box:critic_template}
\end{figure}

\paragraph{Base Models, Computational Environment, and Inference Setup} In this study, we utilized six different models downloaded from HuggingFace Transformers~\footnote{https://huggingface.co/}. We adhered to the licensing terms of all involved models. meta-llama/Llama-3.2-3B-Instruct (LLaMA 3B), meta-llama/Llama-3.1-8B-Instruct (LLaMA 8B) from~\cite{llama3}, and Qwen/Qwen2.5-3B-Instruct (Qwen 3B), Qwen/Qwen2.5-7B-Instruct (Qwen 7B), Qwen/Qwen2.5-14B-Instruct (Qwen 14B), Qwen/Qwen2.5-32B-Instruct (Qwen 32B) from~\cite{qwen2.5,qwen2_5_report}. 

All generative models were trained using either 4 $\times$ A100 80G or 4 $\times$ H100 GPUs. 

To ensure reproducibility, all evaluations are done using zero-shot prompting with greedy decoding and a temperature of 0. Inference of LLMs is carried out using vLLM~\cite{kwon2023efficient}. We utilized the same prompt templates and score extractor as released by~\cite{li_emnlp2024}. Prompt templates for ASAP 1 (Figure \ref{box:error_1}), ASAP 2 (Figure \ref{box:correct_2}), ASAP 5 (Figure \ref{box:main_example}), and ASAP 6 (Figure \ref{box:error_gpt4o}) can also be found in each case studies.

\paragraph{Manual Evaluation Setup} We randomly sampled 20 instances from each dataset and manually examined the reflection and refinement generated. The outputs were derived from a single run using the LLaMA 3B Reasoner and LLaMA 3B Critic model, as reported in Table \ref{tab:main_assessment_comparison}. The annotations were conducted by the authors of this paper. We categorized the errors using the following schema.

\paragraph{Evaluation on Critic's Reflection} Errors in the Critic model's reflections were classified as follows:
\begin{itemize}[leftmargin=*,noitemsep,topsep=0pt]
\item \textbf{Correct Reflection}: The Critic model accurately identified errors in the previous assessment, ensuring faithfulness to both the student's answer and the question content.
\item \textbf{Incorrect Reflection}: The Critic model either misinterpreted the meaning of the student's answer or the scope of key answer elements, leading to incorrect identification of errors or the identification of errors that were not coherent to the given content.
\end{itemize}

\paragraph{Evaluation on Reasoner's Refinement} We classify the error made by the Reasoner model in refinement into the following three categories:
\begin{itemize}[leftmargin=*,noitemsep,topsep=0pt]
\item \textbf{Correct Refinement}: The situation the Reasoner model successfully refined its previous mistakes based on the Critic's reflection.
\item \textbf{Wrong Refinement Obeyed Reflection}: The situation Reasoner model made an error because it faithfully followed the Critic's wrong reflection.
\item \textbf{Wrong Refinement Ignored Reflection}: The situation in which the Reasoner model introduced a new error, deviating from the Critic's reflection.
\end{itemize}

%% file: sections/7_further_exps.tex
\section{Further Experiment Result} \label{sec:further_experiment}

\subsection{Explanation for Main Example} \label{sec:figure_full_example}
As illustrated in Figure \ref{box:main_example}, we present the complete example corresponding to Figure \ref{fig:framework}. 

Initially, the Reasoner takes the question prompt as input and generates its first assessment decision \textcircled{\raisebox{-0.3pt} {\scriptsize2}}. However, in this first attempt, the model incorrectly evaluates the student’s response by crediting key elements such as ``…described mRNA exiting the nucleus…'' and ``…the corresponding amino acids on tRNA being bonded, and the continuation of amino acid linkage until a stop codon is reached,…'' which were not explicitly mentioned.

The Critic model then takes both the question prompt \textcircled{\raisebox{-0.3pt} {\scriptsize1}} and the Reasoner’s initial assessment \textcircled{\raisebox{-0.3pt} {\scriptsize2}} as input to generate a reflection instruction \textcircled{\raisebox{-0.3pt} {\scriptsize3}}. The Critic accurately identifies the Reasoner’s misjudgment, stating: ``You credited the student for mentioning that the `corresponding amino acids on tRNA are bonded to adjacent tRNA's amino acids' and that `amino acids continue to be linked until a STOP codon is read on the mRNA.' However, upon reviewing the student's response, these elements were not explicitly covered.'' The Critic further instructs the Reasoner to ``Please revisit the student's answer and your rationale, considering these points, and try to generate a more precise assessment that reflects the actual content of the student’s response.''

Subsequently, the Reasoner incorporates the chat history and the Critic's feedback (\textcircled{\raisebox{-0.3pt} {\scriptsize1}}, \textcircled{\raisebox{-0.3pt} {\scriptsize2}}, \textcircled{\raisebox{-0.3pt} {\scriptsize3}}) as input to generate a revised assessment decision. The newly generated Reasoner output \textcircled{\raisebox{-0.3pt} {\scriptsize4}} accurately identifies the key elements in the student's response and corrects the final score assessment.

Finally, the Critic evaluates the updated assessment and generates a termination token, ``[STOP],'' indicating the end of the reasoning loop. This process demonstrates the iterative refinement capability of the proposed dual-model framework, ensuring accurate and explainable assessment evaluations.

\begin{figure*}
\begin{tcolorbox}[
    colback=gray!10,      %
    colframe=gray!80,     %
    title=Case Study 1,
    fonttitle=\bfseries,  %
    rounded corners,
    boxrule=0.5mm,        %
    width=\linewidth
]
\scriptsize
\textcircled{\raisebox{-0.3pt} {\scriptsize1}}\textbf{\emph{[Question Prompt]}:}\\
\text{[Question]}: \\
Starting with mRNA leaving the nucleus, list and describe four major steps involved in protein synthesis. \\\\
\text{[Key Elements]}:\\mRNA exits nucleus via nuclear pore.\\mRNA travels through the cytoplasm to the ribosome or enters the rough endoplasmic reticulum.\\mRNA bases are read in triplets called codons (by rRNA).\\tRNA carrying the complementary (U=A, C+G) anticodon recognizes the complementary codon of the mRNA.\\The corresponding amino acids on the other end of the tRNA are bonded to adjacent tRNA's amino acids.\\A new corresponding amino acid is added to the tRNA.\\Amino acids are linked together to make a protein beginning with a START codon in the P site (initiation).\\Amino acids continue to be linked until a STOP codon is read on the mRNA in the A site (elongation and termination).\\\\\text{[Marking Rubric]}:\\3 points: Four key elements;\\2 points: Three key elements;\\1 point: One or two key elements;\\0 points: Other.\\\textbf{[Student Answer]}: the mRNA, Is met by tRNA with anticodons, that link to the mRNA's codons. This tRNA carries an amino acid that goes with the mRNA's codon. This is done multiple times until there is a stranhd of amino acids. Ribosomes attach these amino acids together.\\Critically assess the \text{[Student Answer]} in light of the \text{[Key Answer Elements]}. Offer a detailed explanation for the score assigned, referencing specific criteria from the \text{[Marking Rubric]}:\\ 
\textcircled{\raisebox{-0.3pt} {\scriptsize2}}\textbf{\emph{[Reasoner]}:}\\
The student's answer scored 3 points according to the marking rubric, which awards 3 points for addressing four key elements. The student successfully described mRNA exiting the nucleus, tRNA carrying complementary anticodons recognizing the mRNA codons, the corresponding amino acids on tRNA being bonded, and the continuation of amino acid linkage until a stop codon is reached, as indicated by '...tRNA with anticodons, that link to the mRNA's codons.' and '...until there is a strand of amino acids. Ribosomes attach these amino acids together.' However, the answer lacked explicit mention of mRNA traveling to the ribosome, reading of mRNA bases in triplets by rRNA, addition of new amino acids to tRNA, and the initiation process at the START codon.\\
\textcircled{\raisebox{-0.3pt} {\scriptsize3}}\textbf{\emph{[Critic]}:}\\
It appears there was a misunderstanding in your assessment of the student's answer. You credited the student for mentioning that the 'corresponding amino acids on tRNA are bonded to adjacent tRNA's amino acids' and that 'amino acids continue to be linked until a STOP codon is read on the mRNA.' However, upon reviewing the student's response, these elements were not explicitly covered. The student's description primarily focused on the interaction between mRNA and tRNA and the eventual formation of a peptide bond, but did not detail the process of amino acids being bonded to adjacent tRNA's amino acids nor the specific mechanism of elongation and termination involving a STOP codon. It's crucial to align your assessment closely with the content explicitly presented by the student to ensure accurate scoring. Reflecting on this, you might want to re-evaluate how you interpret and assess answers that do not explicitly cover all the elements listed in a rubric. Please revisit the student's answer and your rationale, considering these points, and try to generate a more precise assessment that reflects the actual content of the student's response.\\
\textcircled{\raisebox{-0.3pt} {\scriptsize4}}\textbf{\emph{[Reasoner]}:}\\
The student's answer scored 2 points according to the marking rubric which stipulates 2 points for achieving three key answer elements. The student accurately described the roles of mRNA and tRNA in protein synthesis, mentioning that 'the mRNA, Is met by tRNA with anticodons, that link to the mRNA's codons' which covers the key elements of mRNA bases being read in triplets and tRNA carrying the complementary anticodon recognizing the complementary codon of the mRNA. Additionally, the student implied the bonding of amino acids by stating 'Ribosomes attach these amino acids together.' However, the answer did not explicitly mention the exit of mRNA from the nucleus, the travel of mRNA through the cytoplasm, the addition of new amino acids to tRNA, the initiation process with a START codon, or the elongation and termination at a STOP codon which resulted in losing points for these key elements.\\
\textcircled{\raisebox{-0.3pt} {\scriptsize5}}\textbf{\emph{[Critic]}:}\\
 Rationale Looks Good! [STOP]
\end{tcolorbox}
\caption{\textbf{The full example as presented in Figure \ref{fig:framework}.}}
\label{box:main_example}
\end{figure*}

\subsection{Case Studies on GPT-4-turbo as Critic} \label{sec:gpt4o_case_study}
The case study in Figure \ref{box:error_gpt4o} highlights the limitations of using GPT-4-turbo as a Critic model. GPT-4-turbo generated feedback tends to be vague, overemphasizing surface-level details while lacking contextual relevance and actionable insights. It struggles to provide precise guidance for improving assessments, often failing to align with key rubric elements and offering inconsistent or generalized reflection instructions. Specifically, the original Reasoner's assessment is correct, but the GPT-4-turbo fails to evaluate the assessment and didn't terminate the iterative refinement process. These shortcomings hinder its effectiveness in refining assessment rationales, underscoring the need for a more tailored Critic model that delivers targeted, domain-specific feedback for accurate and meaningful evaluation.

\begin{figure*}
\begin{tcolorbox}[
    colback=gray!10,      %
    colframe=gray!80,     %
    title=Case Study 2,
    fonttitle=\bfseries,  %
    rounded corners,
    boxrule=0.5mm,        %
    width=\linewidth
]
\scriptsize
\textcircled{\raisebox{-0.3pt} {\scriptsize1}}\textbf{\emph{[Question Prompt]}:}\\
\text{[Question]}: \\
List and describe three processes used by cells to control the movement of substances across the cell membrane.\\
\text{[Key elements]}:\\
Selective permeability is used by the cell membrane to allow certain substances to move across.\\
Passive transport occurs when substances move from an area of higher concentration to an area of lower\\
concentration.\\
Osmosis is the diffusion of water across the cell membrane.\\
Facilitated diffusion occurs when the membrane controls the pathway for a particle to enter or leave a cell.\\
Active transport occurs when a cell uses energy to move a substance across the cell membrane, and\/or a\\
substance moves from an area of low to high concentration, or against the concentration gradient.\\
Pumps are used to move charged particles like sodium and potassium ions through membranes using energy and\\
carrier proteins.\\
Membrane-assisted transport occurs when the membrane of the vesicle fuses with the cell membrane forcing large\\
molecules out of the cell as in exocytosis.\\
Membrane-assisted transport occurs when molecules are engulfed by the cell membrane as in endocytosis.\\
Membrane-assisted transport occurs when vesicles are formed around large molecules as in phagocytosis.\\
Membrane-assisted transport occurs when vesicles are formed around liquid droplets as in pinocytosis.\\
Protein channels or channel proteins allow for the movement of specific molecules or substances into or out of\\
the cell.\\
\text{[Marking Rubric]}:\\
3 points: Three key elements;\\
2 points: Two key elements;\\
1 point: One key element;\\
0 points: Other.\\
\textbf{[Student Answer]}: Three processes used by cells to control the movement of substances across the cell membrane are. * The golgi bodies  * lysosomes * protozoas\\
Evaluate the \text{[Student Answer]} based on the provided \text{[Key Answer Elements]} and \text{[Marking Rubric]}. Summarize the assessment and justify the score awarded:\\
\textcircled{\raisebox{-0.3pt} {\scriptsize2}}\textbf{\emph{[Reasoner]}:}\\
The student's answer scored 0 points according to the marking rubric that awards 3 points for three key elements, 2 points for two key elements, 1 point for one key element, and 0 points for other. The student's response listed 'The golgi bodies, lysosomes, protozoas' which do not pertain to any of the key elements required for the question about processes used by cells to control the movement of substances across the cell membrane. None of the listed items describe selective permeability, passive transport, osmosis, facilitated diffusion, active transport, the use of pumps, or any form of membrane-assisted transport such as exocytosis, endocytosis, phagocytosis, pinocytosis, or the function of protein channels or channel proteins.\\
\textcircled{\raisebox{-0.3pt} {\scriptsize3}}\textbf{\emph{[Critic]}:}\\
 The student's answer does not align with the key elements required to address the question effectively. Here is constructive feedback to help improve their response:\\
1. **Highlight Areas for Improvement**:  \\
   - The response does not mention any relevant cellular processes used to control the movement of substances across the cell membrane, such as selective permeability, passive transport, active transport, or membrane-assisted transport.  \\
   - The listed terms ('Golgi bodies,' 'lysosomes,' 'protozoas') do not pertain to the question and indicate a lack of understanding of the topic.\\
   2. **Provide Actionable Suggestions**:  \\
   - Review the concept of the cell membrane and its role in controlling the movement of substances. Focus on understanding key processes such as passive transport (including osmosis and facilitated diffusion), active transport, and membrane-assisted transport mechanisms like endocytosis and exocytosis.  \\
   - Refer to class materials, textbooks, or trusted educational resources to study how these processes work and why they are important for cellular function.  \\
   - Practice answering similar questions by first identifying the processes relevant to the question and then providing clear descriptions of how each process works. For example, explain that passive transport involves movement along a concentration gradient without energy use, while active transport requires energy to move substances against their concentration gradient.  \\
   - Avoid listing unrelated terms and ensure that each term directly addresses the question.  \\
   By following these steps, you can develop a more accurate and complete response to similar questions in the future. Keep practicing to strengthen your understanding of cell membrane transport mechanisms!\\
\textcircled{\raisebox{-0.3pt} {\scriptsize4}}\textbf{\emph{[Reasoner]}:}\\
To improve the answer, the student should focus on describing specific processes such as passive transport, active transport, and membrane-assisted transport. For example, they could explain how osmosis involves the movement of water across a semi-permeable membrane from an area of lower solute concentration to an area of higher solute concentration, or how active transport requires energy to move substances against their concentration gradient. Additionally, describing the roles of protein channels and carrier proteins in facilitated diffusion and the process of endocytosis and exocytosis would also be relevant.\\
...
\end{tcolorbox}
\caption{\textbf{Prompting GPT-4-turbo failed to act as effective critic model.}}
\label{box:error_gpt4o}
\end{figure*}

\begin{figure*}[!t]
\centering
\includegraphics[width=\linewidth]{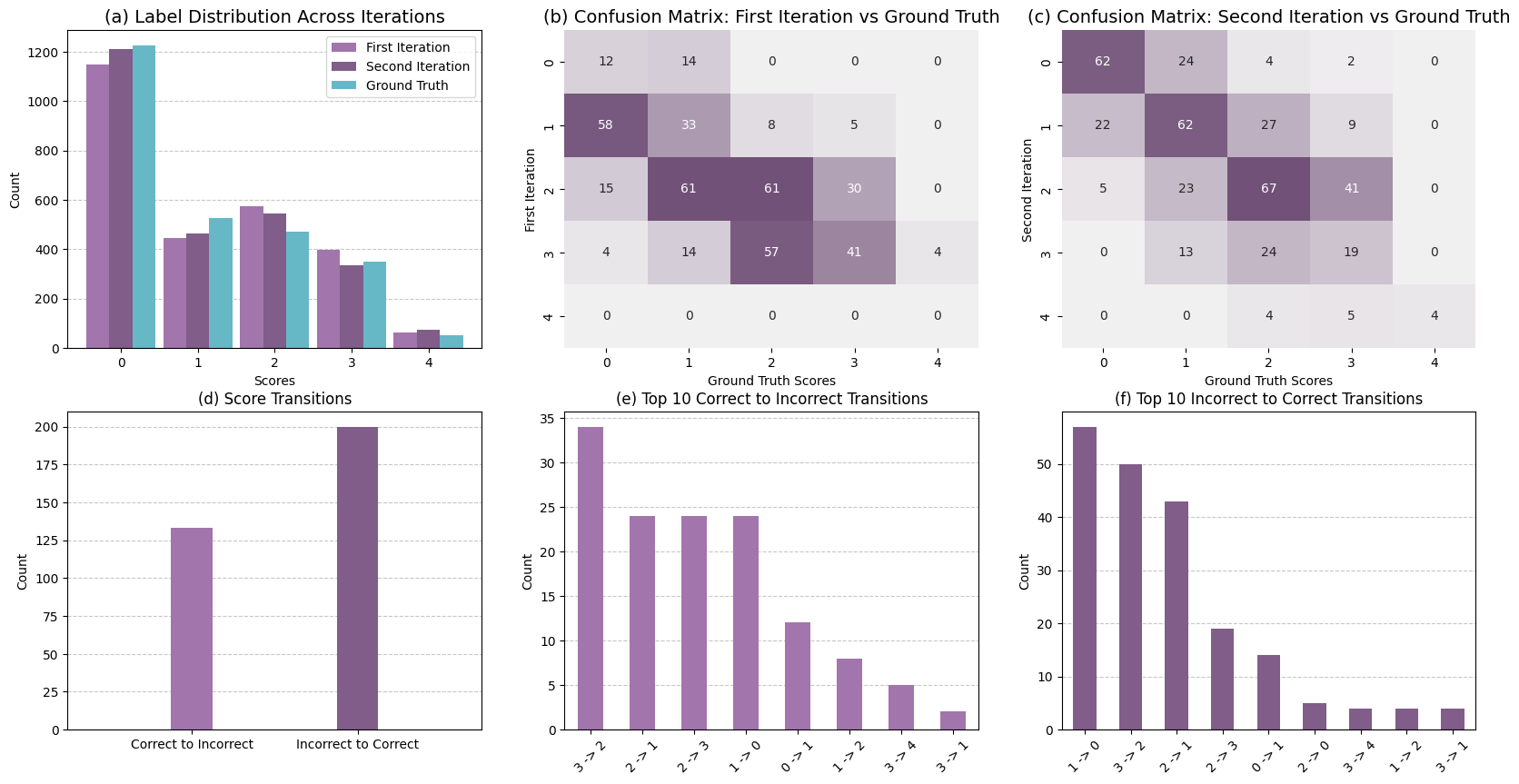}
\caption{\small \textbf{Visualization of detailed error analysis for the iterative reasoning process.}}
\label{fig:further_error_analysis}
\end{figure*}

\subsection{Detailed Error Analysis}  \label{sec:our_detail_error_analysis}

As shown in Figure \ref{fig:further_error_analysis}, we provide an in-depth analysis of the Critic model's effectiveness using a single run with the LLaMA 3B Reasoner and LLaMA 3B Critic model.

\paragraph{Label Distribution} The first row of the Figure \ref{fig:further_error_analysis} presents an analysis of the overall label distribution changes across iterations. As shown in (a), the label distribution shifts closer to the ground-truth distribution after the second iteration with the Critic model's guidance. This trend is further supported by the confusion matrices in (b) and (c), where the second iteration exhibits a more pronounced diagonal pattern, indicating improved alignment with ground-truth labels. In contrast, the first iteration shows a bias towards scores of 0 and 1.

\paragraph{Score Transitions} To gain deeper insights into label transitions, the second row of the Figure \ref{fig:further_error_analysis} examines label changes across iterations. As shown in (d), while our framework does not guarantee perfect label corrections, the majority of transitions move from incorrect to correct labels. This underscores the potential to further refine the collaboration between the Critic and Reasoner models to minimize cases where correct predictions are mistakenly altered. Additionally, (e) and (f) display the top 10 transitions from correct to incorrect and incorrect to correct labels, respectively. The results reveal that most label changes occur between scores of 1 and 3, with the majority involving a single-point difference, reflecting patterns observed in human assessment behaviour.

\subsection{Two Smaller Models May Better Than a Larger One}
\begin{figure}[ht]
\centering
\includegraphics[width=\linewidth]{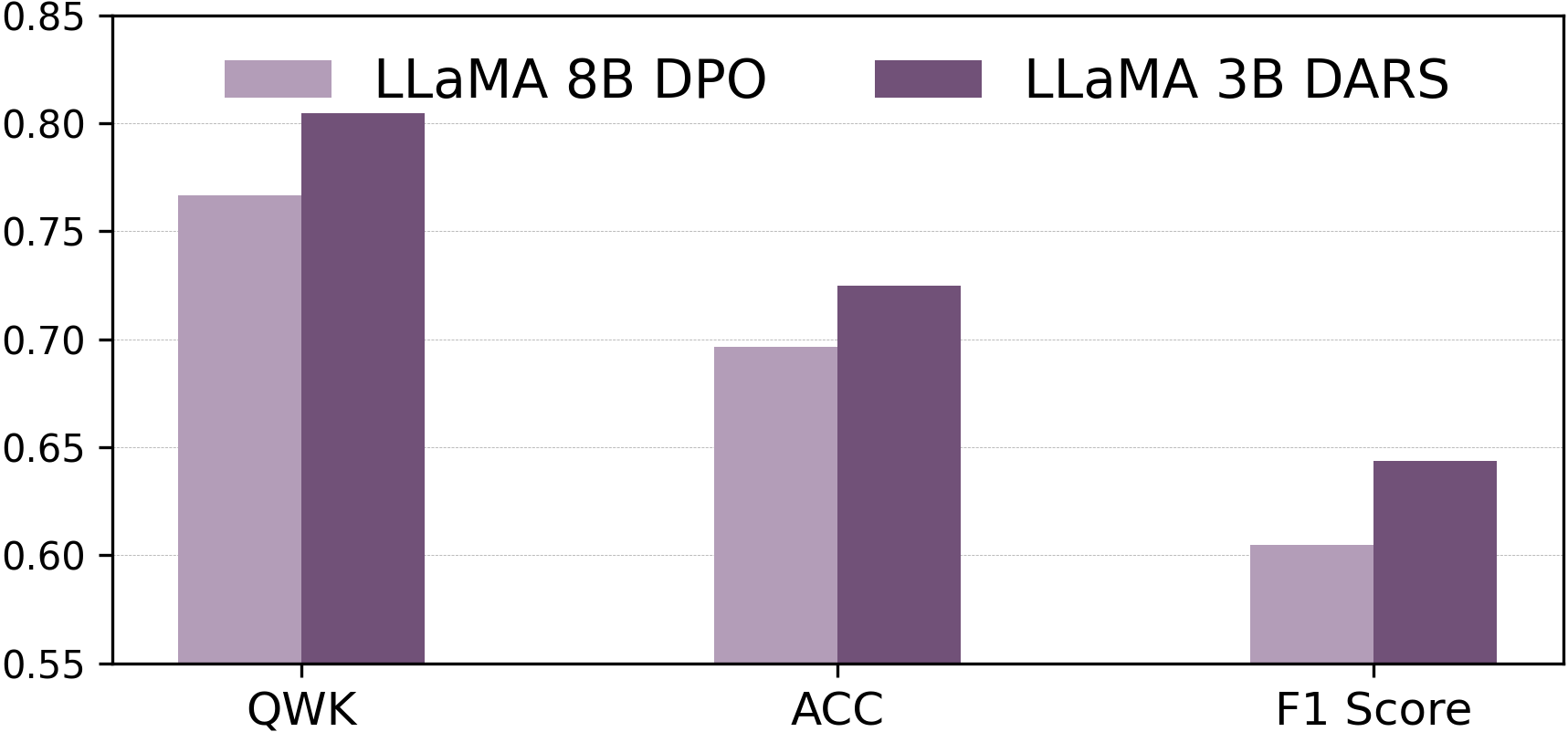}
\caption{\textbf{Comparison of \texttt{DARS} with LLaMA 8B DPO.}}
\label{fig:llama8b_dpo}
\end{figure}

As illustrated in Figure \ref{fig:llama8b_dpo}, \texttt{DARS}, which employs a dual-model setup with LLaMA 3B Reasoner and Critic, outperforms a single LLaMA 8B DPO model. This finding further reinforces that ``two heads are better than one'', demonstrating that two smaller 3B models working together can achieve better results than a single, larger 8B Reasoner. This superior performance may be due to the fact that LLaMA 3B is a distilled variant of the 8B version~\cite{llama3}.  

\subsection{Can Refinement Data Enhance Preference Optimization for the Reasoner?} \label{sec:rationale_dpo}
\begin{figure}[!h]
\centering
\includegraphics[width=\linewidth]{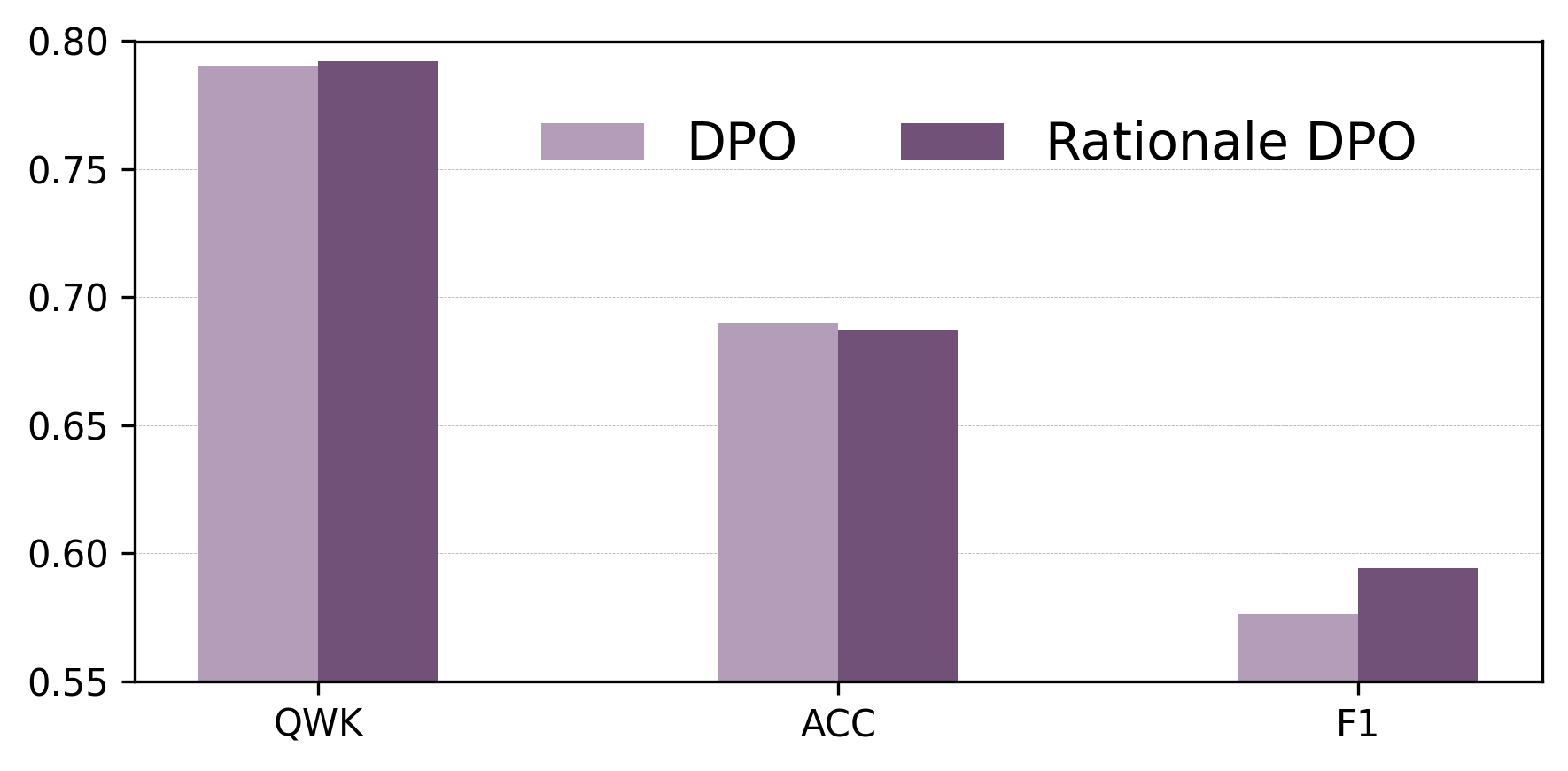}
\caption{\textbf{Regulating DPO training with generated reflections.}}
\label{fig:rationale_po}
\end{figure}Inspired by~\cite{Ref_PO}, we propose a robust preference optimization baseline by incorporating an additional SFT loss on the synthetic reflection data to regularize the DPO training process. As illustrated in Figure \ref{fig:rationale_po}, the inclusion of regularization on reflection data leads to slight improvements in QWK and F1 scores compared with vanilla DPO. These results suggest that \textbf{\emph{refinement data can also serve as an effective regularizer even for single-reasoner training methods}}, enhancing both performance and stability during preference optimisation.

\subsection{Case Studies on Our Framework} \label{sec:our_case_study}

\paragraph{Critic Oversees Errors and Misinterpret Scopes}
As shown in Figure \ref{box:error_1}, the correct assessment of the student’s answer is actually \textbf{1 point}, not 2 or 3. Although the student lists three items, the first item (volume of vinegar) cleanly maps to the ``additional information'' that is missing from the procedure. The other two points are either too vague or already addressed in the procedure (e.g., ``Determine the mass of each sample'' is mentioned, and the procedure does not necessarily require the exact measuring method). Therefore, the response only provides one distinct piece of new information that truly helps replicate the experiment.

The reasoner miscounted the distinct, missing details in the student’s answer. The critic model fails to point this oversee. Although three items were listed—vinegar volume, distilled water volume, and mass measurement method—only one (the amount of vinegar) was truly new. The other two were too vague or already in the procedure, leading the reasoner to mistakenly award 2 and 3 points instead of the correct score of 1.

\paragraph{Critic Correctly Identify Intermediate Errors Even Final Scores are Correct}
As shown in Figure \ref{box:correct_2}, the ``reasoner'' ultimately awarded the correct score of 2 points but incorrectly characterized the student's conclusion as valid. The ``critic'' accurately identified that while the conclusion (``plastic C will take the most weight'') was not supported by the data, the student still described two valid improvements (more trials, ensuring uniform sample length). This discrepancy shows that the critic model can detect errors in the reasoning—namely, that the conclusion is wrong—even when the final numerical score is correct for other reasons (i.e., providing two legitimate design improvements).

\begin{figure*}
\begin{tcolorbox}[
    colback=gray!10,      %
    colframe=gray!80,     %
    title=Case Study 3,
    fonttitle=\bfseries,  %
    rounded corners,
    boxrule=0.5mm,        %
    width=\linewidth
]
\scriptsize
\textcircled{\raisebox{-0.3pt} {\scriptsize1}}\textbf{\emph{[Question Prompt]}:}\\
\text{[Question]}: \\
A group of students wrote the following procedure for their investigation.\\
Procedure:\\
1.Determine the mass of four different samples.\\
2.Pour vinegar in each of four separate, but identical, containers.\\
3.Place a sample of one material into one container and label. Repeat with remaining samples, placing a single sample into a single container.\\
4.After 24 hours, remove the samples from the containers and rinse each sample with distilled water.\\
5.Allow the samples to sit and dry for 30 minutes.\\
6.Determine the mass of each sample.\\
The students's data are recorded in the table below.\\
A table contains four columns: Sample, Starting Mass (g), Ending Mass (g), Difference in Mass (g).\\
The sample for the first row is Marble, with 9.8 Starting Mass, 9.4 Ending Mass and -0.4 for Difference in Mass.\\
The sample for the second row is Limestone, with 10.4 Starting Mass, 9.1 Ending Mass and -1.3 for Difference in Mass.\\
The sample for the third row is Wood, with 11.2 Starting Mass, 11.2 Ending Mass and 0.0 for Difference in Mass.\\
The sample for last row is Plastic, with 7.2 Starting Mass, 7.1 Ending Mass and -0.1 for Difference in Mass.\\
After reading the group's procedure, describe what additional information you would need in order to replicate the experiment.\\
Make sure to include at least three pieces of information.\\
\text{[Key Elements]}: \\
Needed Information:\\
You need to know how much vinegar was used in each container.\\
You need to know what type of vinegar was used in each container.\\
You need to know what materials to test.\\
You need to know what size/surface area of materials should be used.\\
You need to know how long each sample was rinsed in distilled water.\\
You need to know what drying method to use.\\
You need to know what size/type of container to use.\\
Other acceptable responses.\\
\text{[Marking Rubric]}:\\
3 points: The response describes three additional pieces of information that would be needed to accurately replicate the experiment;\\
2 points: The response describes two additional pieces of information that would be needed to accurately replicate the experiment;\\
1 point: The response describes one additional piece of information that would be needed to accurately replicate the experiment;\\
0 point: The response describes little or no accurate or relevant information from the acid rain investigation.\\
 \textbf{[Student Answer]:} In order to replicate this experiment, you would need to know how many ml of vinegar they used, how much distilled water to rinse the samples with and how they obtained the mass of each sample.\\
 Analyze the \text{[Student Answer]} thoroughly. Generate a detailed rationale that explains the strengths and weaknesses of the response: \\
\textcircled{\raisebox{-0.3pt} {\scriptsize2}}\textbf{\emph{[Reasoner]}:}\\
 The student's answer scored \textcolor{red}{2 points according to the marking rubric, as it describes two additional pieces of information needed to replicate the experiment}: the amount of vinegar used ('how many ml of vinegar they used') and the amount of distilled water used for rinsing ('how much distilled water to rinse the samples with'). The student did not specify other necessary details such as the type of vinegar, the materials to test, the size/surface area of materials, the rinsing duration, the drying time, the drying method, or the size/type of container. These missing details are crucial for precise replication and were assessed according to the key answer elements.\\
\textcircled{\raisebox{-0.3pt} {\scriptsize3}}\textbf{\emph{[Critic]}:}\\
 It appears there was a misunderstanding in your assessment of the student's answer. \textcolor{red}{You noted that the student failed to mention the materials to be tested. However, the student did specify the need to know the amount of vinegar and the amount of distilled water used, which indirectly addresses the variety of materials involved in the experiment}. It's crucial to recognize that while the student's response might not explicitly list all desired materials, it does touch upon the conditions necessary to conduct the experiment with various materials. To improve your assessment rationale, focus on what the student has included and ensure that your evaluation aligns with the information provided. Reflect on how each element of the student's answer relates to the experiment's replication and assess accordingly. Please revise your rationale based on this feedback and try to generate a more accurate assessment that acknowledges the student's contributions and missing elements effectively.\\
\textcircled{\raisebox{-0.3pt} {\scriptsize4}}\textbf{\emph{[Reasoner]}:}\\
 The student's answer scored \textcolor{red}{3 points according to the marking rubric, as it describes three additional pieces of information needed to replicate the experiment}: the amount of vinegar ('how many ml of vinegar they used'), the amount of distilled water for rinsing ('how much distilled water to rinse the samples with'), and the drying time ('Allow the samples to sit and dry for 30 minutes'). These details address the key answer elements regarding the amount of vinegar, the materials to test, and the drying time, fulfilling the requirements for a score of 3 points.\\
\textcircled{\raisebox{-0.3pt} {\scriptsize5}}\textbf{\emph{[Critic]}:}\\
 Rationale Looks Good! [STOP]
\end{tcolorbox}
\caption{\textbf{An example that both Reasoner and Critic oversee the mistakes.}}
\label{box:error_1}
\end{figure*}

\begin{figure*}
\begin{tcolorbox}[
    colback=gray!10,      %
    colframe=gray!80,     %
    title=Case Study 4,
    fonttitle=\bfseries,  %
    rounded corners,
    boxrule=0.5mm,        %
    width=\linewidth
]
\scriptsize
\textcircled{\raisebox{-0.3pt} {\scriptsize1}}\textbf{\emph{[Question Prompt]}:}\\
\text{[Question]}: \\
A student performed the following investigation to test four different polymer plastics for stretchability.\\
Procedure:\\
1. Take a sample of one type of plastic, and measure its length.\\
2. Tape the top edge of the plastic sample to a table so that it is hanging freely down the side of the table.\\
3. Attach a clamp to the bottom edge of the plastic sample.\\
4. Add weights to the clamp and allow them to hang for five minutes.\\
5. Remove the weights and clamp, and measure the length of the plastic types.\\
6. Repeat the procedure exactly for the remaining three plastic samples.\\
7. Perform a second trial (T2) exactly like the first trial (T1).\\
The student recorded the following data from the investigation.\\
The table shows the amount of stretch (in millimeters) for four different types of plastic, labeled as A, B,\\
C, and D, when subjected to two different stretching forces, labeled as T1 and T2.\\
For plastic type A, it stretched 10mm under T1 and 12mm under T2.\\
For plastic type B, it stretched 22mm under T1 and 23mm under T2.\\
For plastic type C, it stretched 14mm under T1 and 13mm under T2.\\
Lastly, for plastic type D, it stretched 20mm under both T1 and T2.\\
a. Draw a conclusion based on the student's data.\\
b. Describe two ways the student could have improved the experimental design and\/or validity of the results.\\
\text{[Key Elements]}:\\
Conclusions:\\
Plastic sample B has more stretchability than the other polymer plastics.\\
Plastic sample A has the least amount of stretchability compared to the other polymer plastics.\\
Not all polymer plastics have the same stretchability.\\
Different polymer plastics have different stretchability (and are therefore suited for different applications).\\
A reasonable conclusion cannot be drawn due to procedural errors.\\
Other reasonable conclusions \\
Experimental Design Improvements:\\
Provide the before and after measurements for length (Did the samples all start out the same size?).\\
Make sure the samples are all of the same thickness.\\
Variations in thickness could have caused variations in stretchability.\\
Perform additional trials.\\
Some of the samples have similar stretchability (A and C, B and D).\\
Two trials may not be enough to conclusively state that one is more stretchable than the other.\\
Indicate how many weights were added to the clamps (Was it the same number for each sample?).\\
Other acceptable responses\\
\text{[Marking Rubric]}:\\
3 points: The response draws a valid conclusion supported by the student's data and describes two ways the student could have improved the experimental design and\/or the validity of the results;\\
\textcolor{red}{2 points}: The response draws \textcolor{red}{a valid conclusion} supported by the student's data and describes \textcolor{red}{one way the student could have improved the experimental design} and/or the validity of the results. \textcolor{red}{-or-} The response describes \textcolor{red}{two ways the student could have improved the experimental design} and/or the validity of the results but fails to draw or incorrectly draws a conclusion from the student's data;\\
1 point: The response draws a valid conclusion supported by the student's data but fails to describe, or incorrectly describes, how the student could have improved the experimental design and/or the validity of the results. -or- The response describes one way the student could have improved the experimental design and/or the validity of the results but fails to draw or incorrectly draws a conclusion from the student's data.;\\
0 points: The response provides little or no correct information from the polymer investigation.\\
 \textbf{[Student Answer]:} A. Based on the results of the experiment, I believe that plastic C will take the most weight. All the other plastics either extended when more weight was added or stayed the same. Plastic C though actually shrunk in size unlike plastic A which went from 10 to 12 and B which went from 22-23 mm and lastly D witch remained at 20 mm the whole duration. B: A way the student could of improved the experiment validity is by having more trials and having the plastic strips the same length.\\
 Critically assess the \text{[Student Answer]} in light of the [Key Answer Elements]. Offer a detailed explanation for the score assigned, referencing specific criteria from the [Marking Rubric]: \\
\textcircled{\raisebox{-0.3pt} {\scriptsize2}}\textbf{\emph{[Reasoner]}:}\\
The student's answer scored \textcolor{red}{2 points according to the marking rubric}. The student drew \textcolor{red}{a valid conclusion} supported by the data, stating 'plastic C will take the most weight,' which aligns with the key answer element that a valid experimental conclusion should be drawn. Additionally, the student proposed \textcolor{red}{one valid experimental design improvement} by suggesting 'having more trials,' which aligns with the key answer element about performing additional trials. However, the student did not address other potential improvements such as ensuring uniform sample thickness or specifying the amount of weight added, which limited the score to 2 points.\\
\textcircled{\raisebox{-0.3pt} {\scriptsize3}}\textbf{\emph{[Critic]}:}\\
 It appears there was a misunderstanding in your assessment of the student's answer. \textcolor{red}{The student did not address the experimental conclusion requirements accurately}. The expected conclusions were about the stretchability of different polymer plastics and their suitability for different applications, along with a statement about procedural errors affecting the validity of conclusions. The student focused primarily on which plastic would take the most weight, which is not the correct focal point for the experimental conclusion as per the given rubric. Additionally, while you correctly noted the suggestion for increasing the number of trials as a potential improvement, the student's other suggestions such as 'having the plastic strips the same length' do not directly address more critical factors like uniform thickness or the amount of weight added, which are essential for ensuring experimental validity. When revising your rationale, it would be beneficial to closely align your assessment with the specific requirements outlined in the rubric. Reflect on the key elements missed in the student's response, such as the failure to discuss uniform sample thickness or the amount of weight added, and consider how these omissions affect the score. Please take this feedback into account and attempt to regenerate a more accurate and comprehensive rationale.\\
\textcircled{\raisebox{-0.3pt} {\scriptsize4}}\textbf{\emph{[Reasoner]}:}\\
 The student's answer scores \textcolor{red}{2 points according to the marking rubric}. The student \textcolor{red}{fails to draw a valid conclusion} from the data, incorrectly stating 'plastic C will take the most weight,' while the data shows plastic B stretched the most. However, the student \textcolor{red}{correctly suggests two experimental design improvements}: increasing the number of trials ('...having more trials...') and ensuring uniformity in the length of the plastic samples ('having the plastic strips the same length'). The response thus meets the criteria for 2 points, as it describes two valid ways to improve the experimental design but incorrectly draws a conclusion from the data.\\
\textcircled{\raisebox{-0.3pt} {\scriptsize5}}\textbf{\emph{[Critic]}:}\\
 Rationale Looks Good! [STOP]
\end{tcolorbox}
\caption{\textbf{An example that Critic can identify intermediate errors even final scores are correct.}}
\label{box:correct_2}
\end{figure*}

\subsection{Case Study: Comparing Critic's Output with Different Sizes} \label{sec:compare_critic_sizes}
In Figure~\ref{box:compare_critic_size}, Qwen3B (the reasoner) mistakenly awards the student’s answer 2points rather than the 0 points warranted by the rubric. Comparing critic responses of different model sizes highlights varied degrees of thoroughness in their feedback. Qwen 3B’s critic, for instance, repeatedly instructs that ``\emph{it's important to ensure that the key elements of each biological process are correctly identified and matched ...}'' yet does not provide concrete steps for revising the assessment. By contrast, Qwen7B merely remarks ``\emph{Rationale Looks Good! [STOP]},'' signalling an abrupt end to any meaningful revision. Progressing to Qwen 14B, the critic offers more constructive guidance by urging: ``\emph{You might want to revisit the student’s answer and compare it directly with the key elements required by the rubric.}'' Finally, Qwen 32B delivers the most comprehensive feedback, emphasizing ``\emph{It’s important to verify that the terms and processes described align correctly with biological definitions and mechanisms before awarding points based on the rubric}'' and detailing how the student's descriptions deviate from correct scientific definitions. This gradual increase in clarity, depth, and actionable insights indicates that larger model sizes (14B and 32B) are more effective at diagnosing errors and recommending precise revisions.

\begin{figure*}
\begin{tcolorbox}[
    colback=gray!10,      %
    colframe=gray!80,     %
    title=Case Study 5,
    fonttitle=\bfseries,  %
    rounded corners,
    boxrule=0.5mm,        %
    width=\linewidth
]
\scriptsize
\textcircled{\raisebox{-0.3pt} {\scriptsize1}}\textbf{\emph{[Question Prompt]}:}\\
\text{[Question]}: \\
List and describe three processes used by cells to control the movement of substances across the cell membrane.\\
\text{[Key elements]}:\\
Selective permeability is used by the cell membrane to allow certain substances to move across.\\
Passive transport occurs when substances move from an area of higher concentration to an area of lower\\
concentration.\\
Osmosis is the diffusion of water across the cell membrane.\\
Facilitated diffusion occurs when the membrane controls the pathway for a particle to enter or leave a cell.\\
Active transport occurs when a cell uses energy to move a substance across the cell membrane, and\/or a\\
substance moves from an area of low to high concentration, or against the concentration gradient.\\
Pumps are used to move charged particles like sodium and potassium ions through membranes using energy and\\
carrier proteins.\\
Membrane-assisted transport occurs when the membrane of the vesicle fuses with the cell membrane forcing large\\
molecules out of the cell as in exocytosis.\\
Membrane-assisted transport occurs when molecules are engulfed by the cell membrane as in endocytosis.\\
Membrane-assisted transport occurs when vesicles are formed around large molecules as in phagocytosis.\\
Membrane-assisted transport occurs when vesicles are formed around liquid droplets as in pinocytosis.\\
Protein channels or channel proteins allow for the movement of specific molecules or substances into or out of\\
the cell.\\
\text{[Marking Rubric]}:\\
3 points: Three key elements;\\
2 points: Two key elements;\\
1 point: One key element;\\
0 points: Other.\\
\textbf{[Student Answer]}: Endocyotsis, when food enters the cell through the cell membrane, exocytosis, when food exits as waste out of the cell, and translation, when the cell surrounds the food, and engulfs it.\\
Evaluate the \text{[Student Answer]} based on the provided \text{[Key Answer Elements]} and \text{[Marking Rubric]}. Summarize the assessment and justify the score awarded:\\
\textcircled{\raisebox{-0.3pt} {\scriptsize2}}\textbf{\emph{[Reasoner] Qwen 3B}:}\\\\
The student's answer scored 2 points according to the marking rubric, which awards 2 points for correctly addressing two key elements. The student successfully described 'endocytosis' and 'exocytosis' as processes where 'the cell surrounds the food, and engulfs it' and 'when food enters the cell through the cell membrane, exocytosis, when food exits as waste out of the cell,' respectively. These descriptions align with the key elements of membrane-assisted transport when molecules are engulfed by the cell membrane and vesicles are formed around large molecules, as well as exocytosis. However, the answer lacked descriptions of selective permeability, passive transport, osmosis, facilitated diffusion, active transport, the use of pumps, and protein channels, which were necessary to meet additional key elements for a higher score.\\
\textbf{\emph{[Critic] Qwen 3B}:}\\
In reviewing the assessment rationale you provided, it appears there was a misunderstanding in the evaluation of the student's answer. Specifically, the student's description of 'endocytosis' and 'exocytosis' was misinterpreted in key aspects. The student mentioned that 'the cell surrounds the food, and engulfs it' and 'when food exits as waste out of the cell,' respectively. However, these descriptions do not accurately reflect the scientific definitions and processes of endocytosis and exocytosis. Endocytosis involves the cell membrane engulfing materials to form a vesicle, which is then transported into the cell, often used for nutrient absorption or waste expulsion. Exocytosis, on the other hand, involves the cell membrane expelling materials to form a vesicle, which is then transported out of the cell. The student's answer did not mention the critical aspects of vesicle formation or the directional movement of materials in both processes. It's important to ensure that the key elements of each biological process are correctly identified and matched with the student's descriptions before assigning points based on the marking rubric. This will help in providing accurate feedback and scores that truly reflect the student's understanding of the subject matter. As you reflect on this feedback, please revisit the student's answer and the marking rubric, and consider how each part of the student's response aligns with the expected scientific definitions and processes. After reflecting on these points, I encourage you to revise and regenerate the assessment rationale to more accurately represent the student's understanding and the biological processes in question.\\
\textbf{\emph{[Critic] Qwen 7B}:}\\
Rationale Looks Good! [STOP]\\
\textbf{\emph{[Critic] Qwen 14B}:}\\
It appears there was a misunderstanding in your assessment of the student's answer regarding the key elements of endocytosis and exocytosis. You noted that the student described these processes as involving 'membrane-assisted transport when molecules are engulfed by the cell membrane' and 'when vesicles are formed around large molecules.' However, the student's response did not explicitly mention these mechanisms. Instead, they provided general descriptions of the processes without detailing the specific mechanisms like 'membrane-assisted transport' or 'formation of vesicles around molecules.' It's crucial to align your assessment with the exact content provided by the student to ensure accuracy in scoring. You might want to revisit the student's answer and compare it directly with the key elements required by the rubric. This will help you identify precisely what the student mentioned versus what was omitted. A useful strategy could be to mark the presence of each required element in the answer, noting exactly what terms or concepts are mentioned that correspond to those in the rubric. This exercise will not only help in more accurate scoring but also in providing constructive feedback to the student. Reflect on this approach and try to revise your rationale based on these insights.\\
\textbf{\emph{[Critic] Qwen 32B}:}\\
It appears there was a misunderstanding in the assessment of the student's answer regarding the description of 'endocytosis' and 'exocytosis'. You credited the student for accurately describing these processes in terms of membrane-assisted transport and formation of vesicles around large molecules. However, the student's description was inaccurate and did not clearly distinguish between the processes and concepts involved. The student incorrectly associated 'endocytosis' with the cell surrounding and engulfing food, which is a misrepresentation of the process. Similarly, the mention of 'exocytosis' was coupled with an incorrect description of food exiting the cell, rather than correctly describing it as the process where substances are expelled from the cell through vesicle fusion with the cell membrane. Furthermore, the student's answer included 'translation,' which is not a membrane-assisted transport process but rather a stage of protein synthesis. This indicates a fundamental misunderstanding or misstatement of the biological processes involved. To improve your assessment rationale, you should focus on the accuracy of the student's descriptions in relation to the scientific concepts asked in the question. It's important to verify that the terms and processes described align correctly with biological definitions and mechanisms before awarding points based on the rubric. Reflect on these points and consider how a more accurate assessment can be framed, emphasizing the importance of precise scientific terminology and processes. Please revise your rationale based on this feedback and try to generate a more accurate assessment of the student's answer.
\end{tcolorbox}
\caption{\textbf{Comparing Critic model's output with different parameter sizes.}}
\label{box:compare_critic_size}
\end{figure*}